%% file: example_paper.tex
\documentclass{article}

\usepackage{microtype}
\usepackage{graphicx}
\usepackage{subcaption}
\usepackage{booktabs} % for professional tables

\usepackage{hyperref}

\usepackage[accepted]{icml2026}

\usepackage{amsmath}
\usepackage{amssymb}
\usepackage{mathtools}
\usepackage{amsthm}

\usepackage{hyperref}
\usepackage{url}
\usepackage{xspace}
\usepackage{enumitem}
\usepackage{graphicx}
\usepackage{wrapfig}
\usepackage{fontawesome5}
\usepackage{subcaption}
\usepackage{multirow}
\usepackage{enumitem}
\usepackage{booktabs}
\usepackage{xcolor}
\colorlet{grey}{gray} 
\usepackage{wrapfig}
\usepackage{listings}
\usepackage{xcolor}
\usepackage[table]{xcolor}
\usepackage[capitalize]{cleveref}
\usepackage{marvosym}
\usepackage[most]{tcolorbox}
\usepackage{algorithm}
\usepackage{algorithmic}

\newtcolorbox{promptbox}[1]{
    title=#1,
    colback=white,
    colframe=black,
    boxrule=1pt,
    arc=5pt,
    left=6pt,
    right=6pt,
    top=6pt,
    bottom=6pt,
    fontupper=\normalfont,
    colbacktitle=grey!20!white,
    coltitle=black,
    fonttitle=\bfseries,
    breakable % 允许跨页断行
}
\newcommand{\framework}{\textbf{CPM\"{o}bius}\xspace}
\definecolor{boxframe}{HTML}{7BA7D6} 
\definecolor{boxback}{HTML}{F5F9FF} 
\definecolor{boxtitle}{HTML}{E6F0FF} 
\definecolor{boxtitleline}{HTML}{6A98C9}

\colorlet{punct}{red!60!black}
\definecolor{background}{HTML}{EEEEEE}
\definecolor{delim}{RGB}{20,105,176}
\colorlet{numb}{magenta!60!black}
\theoremstyle{plain}

\theoremstyle{definition}

\theoremstyle{remark}

\newcommand{\TO}{\text{to }}
\newcommand{\RETURN}{\textbf{return }}
\usepackage[textsize=tiny]{todonotes}

\usepackage[capitalize]{cleveref}

\icmltitlerunning{CPM\"{o}bius: Iterative Coach–Player Reasoning for Data-Free Reinforcement Learning}

\begin{document}

\twocolumn[
  \icmltitle{CPM\"{o}bius: Iterative Coach–Player Reasoning \\
  for Data-Free Reinforcement Learning}

  % It is OKAY to include author information, even for blind submissions: the
  % style file will automatically remove it for you unless you've provided
  % the [accepted] option to the icml2026 package.

  % List of affiliations: The first argument should be a (short) identifier you
  % will use later to specify author affiliations Academic affiliations
  % should list Department, University, City, Region, Country Industry
  % affiliations should list Company, City, Region, Country

  % You can specify symbols, otherwise they are numbered in order. Ideally, you
  % should not use this facility. Affiliations will be numbered in order of
  % appearance and this is the preferred way.
  \icmlsetsymbol{equal}{*}

  \begin{icmlauthorlist}
    \icmlauthor{Ran Li}{equal,yyy,xxx}
    \icmlauthor{Zeyuan Liu}{equal,yyy}
    \icmlauthor{Yinghao Chen}{yyy}
    \icmlauthor{Bingxiang He}{yyy}
    \icmlauthor{Jiarui Yuan}{yyy}
    \icmlauthor{Zixuan Fu}{yyy}
    \icmlauthor{Weize Chen}{yyy}
    \icmlauthor{Jinyi Hu}{yyy}
    \icmlauthor{Chen Qian}{zzz}
    \icmlauthor{Zhiyuan Liu}{yyy}
    \icmlauthor{Maosong Sun}{yyy}
  \end{icmlauthorlist}

  % \icmlaffiliation{yyy}{Department of Computer Science and Technology, Tsinghua \\ University, }
  % \icmlaffiliation{xxx}{Department of Computer Science and Technology \\, University of Cambridge}
  \icmlaffiliation{yyy}{Tsinghua University,}
  \icmlaffiliation{xxx}{University of Cambridge}
  \icmlaffiliation{zzz}{Shanghai Jiao Tong University}
  \icmlcorrespondingauthor{Ran Li}{rl810@cam.ac.uk}
  \icmlcorrespondingauthor{Weize Chen}{chenwz21@mails.tsinghua.edu.cn}
  \icmlcorrespondingauthor{Zhiyuan Liu}{liuzy@tsinghua.edu.cn}

  % You may provide any keywords that you find helpful for describing your
  % paper; these are used to populate the "keywords" metadata in the PDF but
  % will not be shown in the document
  \icmlkeywords{Machine Learning, ICML}

  \vskip 0.1in
]

% this must go after the closing bracket ] following \twocolumn[ ...

% This command actually creates the footnote in the first column listing the
% affiliations and the copyright notice. The command takes one argument, which
% is text to display at the start of the footnote. The \icmlEqualContribution
% command is standard text for equal contribution. Remove it (just {}) if you
% do not need this facility.

% Use ONE of the following lines. DO NOT remove the command.
% If you have no special notice, KEEP empty braces:
\printAffiliationsAndNotice{}  % no special notice (required even if empty)
% Or, if applicable, use the standard equal contribution text:
% \printAffiliationsAndNotice{\icmlEqualContribution}

\begin{abstract}
  Large Language Models (LLMs) have demonstrated strong potential in complex reasoning, yet their progress remains fundamentally constrained by reliance on massive high-quality human-curated tasks and labels, either through supervised fine-tuning (SFT) or reinforcement learning (RL) on reasoning-specific data. This dependence renders supervision-heavy training paradigms increasingly unsustainable, with signs of diminishing scalability already evident in practice. To overcome this limitation, we introduce \framework, a collaborative \textbf{Coach–Player} paradigm for data-free reinforcement learning of reasoning models. Unlike traditional adversarial self-play frameworks, \framework inspired by multi-agent collaboration, treats the Coach and Player as independent but cooperative roles. The Coach proposes instructions targeted at the Player’s capability and receives rewards based on changes in the Player’s performance, while the Player is rewarded for solving the increasingly instructive tasks generated by the Coach. This cooperative optimization loop is designed to directly enhance the Player’s mathematical reasoning ability. Remarkably, \framework achieves substantial improvement without relying on any external training data, outperforming existing unsupervised approaches. For example, on Qwen2.5-Math-7B-Instruct, our method improves accuracy by overall average +4.9 and out-of-distribution average +5.4, which exceed RENT by +1.5 on overall accuracy and R-zero by +4.2 on OOD accuracy.  Our codebase has been released at \url{https://github.com/thunlp/CPMobius}. \looseness=-1
\end{abstract}

\input{sections/introduction}
\input{sections/related_work}
\input{sections/Preliminaries}

\input{sections/framework}

\input{sections/experiment}
% \input{sections/ablation}
\input{sections/conclusion}

\section*{Impact Statement}
This work introduces CPMöbius, a data-free reinforcement learning framework that enhances reasoning in large language models through a cooperative Coach–Player paradigm. Because our method does not require human-annotated data or human feedback during training, it avoids risks associated with large-scale human data collection, such as privacy concerns, labor exploitation, or biased supervision. All experiments were conducted on publicly available benchmark datasets (e.g., AMC, AIME, MATH, OlympiadBench), which are widely used in the research community for evaluating mathematical reasoning models. No personally identifiable, sensitive, or private data was used.
Potential societal impacts include both positive applications, such as advancing safe autonomous reasoning systems, and risks, such as misuse for harmful automated problem-solving. We emphasize that CPMöbius is designed to improve verifiable mathematical reasoning, not to generate unverified or harmful content. Nonetheless, as with any reinforcement learning system, safeguards should be considered in future deployments to mitigate unintended misuse.

\bibliography{example_paper}
\bibliographystyle{icml2026}

\newpage
\appendix
\crefname{appendix}{Appendix}{Appendices}
\crefalias{section}{appendix}
\crefalias{subsection}{appendix}
\onecolumn
\input{sections/appendix}

% You can have as much text here as you want. The main body must be at most $8$
% pages long. For the final version, one more page can be added. If you want, you
% can use an appendix like this one.

% The $\mathtt{\backslash onecolumn}$ command above can be kept in place if you
% prefer a one-column appendix, or can be removed if you prefer a two-column
% appendix.  Apart from this possible change, the style (font size, spacing,
% margins, page numbering, etc.) should be kept the same as the main body.
%%%%%%%%%%%%%%%%%%%%%%%%%%%%%%%%%%%%%%%%%%%%%%%%%%%%%%%%%%%%%%%%%%%%%%%%%%%%%%%
%%%%%%%%%%%%%%%%%%%%%%%%%%%%%%%%%%%%%%%%%%%%%%%%%%%%%%%%%%%%%%%%%%%%%%%%%%%%%%%

\end{document}

%% file: sections/introduction.tex
\section{Introduction}
\label{sec:introduction}
Large Language Models (LLMs)~\citep{openai-gpt5, qwen2, llama} have demonstrated remarkable capabilities in complex reasoning tasks, from mathematical reasoning, problem solving~\citep{wei2022chain} to code generation~\citep{chen2021evaluating}. The dominant paradigm for enhancing these abilities involves post-training on domain-specific data, typically through supervised fine-tuning (SFT)~\citep{ouyang2022training, tunstall2023zephyr} followed by reinforcement learning (RL)~\citep{christiano2017deep, ppo}. While effective, these approaches are fundamentally constrained by their reliance on massive, high-quality, human-curated datasets. The scarcity of such expert-produced examples means this highly supervision-dependent paradigm is showing signs of strain, raising concerns about its long-term scalability. \looseness=-1

% This data bottleneck not only incurs significant costs but also conceptually limits a model's potential to what can be explicitly taught by humans. 
\input{fig_code/clarify}

To break free from this dependency, a promising frontier has emerged in data-free learning, where models improve through autonomous interaction. Self-play, a concept inspired by game-playing AI~\citep{silver2017mastering}, has been adapted for LLM reasoning to achieve self-evolving. Recent self-play frameworks in RL ~\citep{huang2025rzeroselfevolvingreasoningllm,zhao2025absolute} have shown that a model can generate its own training data and learn from solving them, entirely removing the need for external datasets. However, these pioneering methods are often built on an adversarial or competitive dynamic, where the model in one role generates challenges to stump another. Such an adversarial setup is prone to instability, collapsing into nonsensical or unlearnable proposed tasks for RL training. \looseness=-1

In this work, we propose \framework, a Coach–Player paradigm for data-free reinforcement learning, inspired by real world human sports collaboration and multi-agent collaboration~\citep{chen2024internet, qian2023communicative}. Instead of casting the Player model as competitors, the Coach is responsible for adapting the task difficulty to the Player's capabilities. \framework treats the Coach and Player models as independent but collaborative partners in a symbiotic learning process. Throughout this paper, \textbf{“data-free”} refers only to the co-evolution stage after Coach–Player collaboration begins, and does not count any one-time model initialization performed beforehand. As shown in \cref{fig:clarify}, the Coach and Player models are optimized through a cooperative loop:

\begin{itemize}[itemsep=2pt, parsep=0pt, topsep=0pt, partopsep=0pt]
    \item The \textbf{Coach} model acts as a curriculum designer, proposing maximally instructive tasks targeted at the Player’s current capability.
    \item The \textbf{Player} model focuses on solving these tasks to enhance its reasoning skills. \looseness=-1
    \item The \textbf{reward signals} for both Coach and Player are designed to foster cooperation. The Coach is rewarded based on the \textbf{environment feedback-induced accuracy gap of the Player}, directly incentivizing it to generate instructions that lead to tangible learning progress. Simultaneously, the Player is rewarded via a \textbf{standard verifiable outcome} for correctly solving tasks provided by the Coach.\looseness=-1
\end{itemize}

A critical ingredient in this paradigm is a \textbf{Coach that can genuinely “teach”}. It must ask constructive, targeted questions and scaffold the Player with tasks that are informative rather than random. In practice, a weak or unskilled Coach tends to generate ambiguous or unhelpful tasks, which makes the feedback noisy and undermines co-evolution.

This collaborative dynamic allows \framework to generate a highly targeted and adaptive curriculum from scratch, tailored specifically to the Player's evolving needs throughout the training process. Our experiments show that this data-free, cooperative approach is not only viable but remarkably effective. Without relying on any external training data during co-evolution, \framework achieves substantial improvements and outperforms existing unsupervised methods. For instance, on the Qwen2.5-Math-7B-Instruct, our method improves accuracy by overall average +4.9 and out-of-distribution average +5.4, a significant leap compared to the +1.5 from RENT, a method of reinforcement learning via entropy minimization ~\citep{prabhudesai2025maximizing} and +4.2 from R-zero. The details of these baseline methods are provided in~\cref{sec:experiment_setup}  These results demonstrate the effectiveness and scalability of collaborative paradigm as a new pathway for advancing mathematical reasoning in LLMs, decoupling their progress from the constraints of human supervision.

%% file: fig_code/clarify.tex
% \begin{figure}{t}
%     % \vspace{0em}
%     % \hfill
%     \includegraphics[\columnwidth]{figure/clarify.png}
%     \vspace{-2em}
%     \caption{\framework starts with the coach proposing tasks of suitable difficulty. The player learns by solving these tasks, then reviews on a predefined validation set. Finally, the coach adjusts the next training plan based on the player’s performance. }
    
%     \label{fig:clarify}
%     \vspace{-0.5em}
% \end{figure}

\begin{figure}[t] % [t] 表示优先将图片放在栏目顶部
    \centering % 让图片和标题居中
    % 将图片宽度设置为当前栏目的宽度 (\columnwidth)
    \includegraphics[width=\columnwidth]{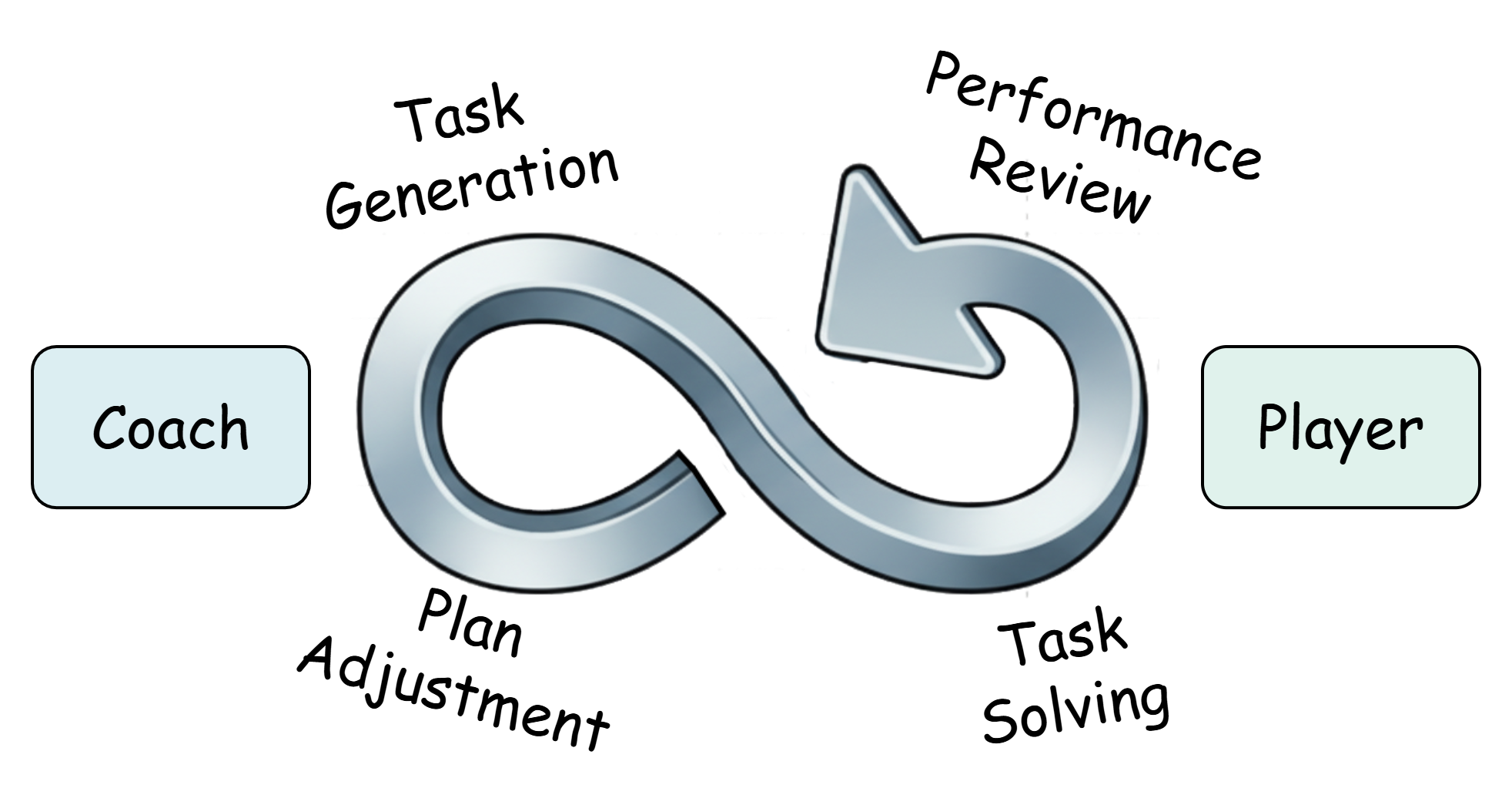}
    
    % 建议先去掉手动的 \vspace，使用默认间距，有需要再微调
    % \vspace{-2em} 
    
    \caption{\framework starts with the coach proposing tasks of suitable difficulty. The player learns by solving these tasks, then reviews on a predefined environment. Finally, the coach adjusts the next training plan based on the player’s performance.}
    \label{fig:clarify}
    
    % \vspace{-0.5em}
\end{figure}

%% file: sections/related_work.tex
\section{Related Work}
\label{sec:related_work}

\paragraph{Reinforcement Learning with Verifiable Rewards.}

Recent advances in language model reasoning have leveraged Reinforcement Learning with Verifiable Rewards (RLVR), in which models are trained using binary feedback derived from programmatic verifiers that check correctness against ground truth~\citep{lambert2024tulu,guo2025deepseek,zhang2025rightquestionhalfanswer}. By replacing learned reward models with rule-based verifiers, RLVR enables reliable optimization and mitigates reward hacking. Leading systems~\citep{jaech2024openai,openai-o3,openai-gpt5,openai2025gpt-oss-120b, comanici2025gemini25, seed2025seed1} demonstrate that RLVR can substantially improve reasoning and problem-solving abilities. Typical rule-based rewards include accuracy checks for deterministic outcomes and format constraints for structured outputs, both of which enhance the reliability and reproducibility of large-scale RL training pipelines. Despite their effectiveness, RLVR is fundamentally limited by the availability of verifiable supervision, which becomes increasingly costly as models surpass human-level expertise in specialized domains~\citep{burns2023weak}.

\paragraph{Self-Play and Co-Evolving Policy-Rewards.}

Self-play has emerged as a powerful paradigm for improving LLMs without relying solely on external supervision. In this approach, a model either generates its own training signals or interacts with a counterpart to refine both policy and reward~\citep{yuan2024self,jiang2025pag}. Techniques include self-rewarding, where a model critiques or corrects its own outputs~\citep{xiong2025self,zhang2025critique,kimiteam2025kimik2openagentic}, and co-optimization, where the policy and a separate reward model are trained jointly to enhance robustness and reduce reward hacking~\citep{zha2025rl,hong2025cooper,lu2025urpo}. By unifying the roles of generator and verifier, self-play enables dynamic adaptation and continuous improvement, offering a scalable alternative to purely supervised or static reward schemes.

\paragraph{Data-Free Reinforcement Learning.} 

To address the limitations of human-generated rewards, recent work has explored data-free RL methods that generate training signals automatically. Some approaches leverage a model’s own outputs or internal states, using consistency, confidence, or self-evaluation to guide learning~\citep{zuo2025ttrl,agarwal2025unreasonable,li2025confidence,yuan2024self}. Others rely on external, automated signals, such as heuristics or the structure of large unlabeled corpora~\citep{dong2025reinforcement,zweiger2025self}. More sophisticated methods combine these ideas, allowing models to generate problems for themselves, evaluate solutions, and iteratively refine both policy and reward~\citep{zhao2025absolute,huang2025rzeroselfevolvingreasoningllm,chen2025self,he2026far}. Together, these data-free approaches provide scalable training for LLMs, enabling self-improvement without human labels, though they remain sensitive to reward misalignment and can exhibit failure modes such as collapse or repetitive behavior.

%% file: sections/Preliminaries.tex
\section{Preliminaries}
\label{sec:preliminaries}
In this section, we briefly review two key RL methods for LLM that are relevant to our framework.
\subsection{Group Relative Policy Optimization}
Group Relative Policy Optimization (GRPO)~\citep{shao2024deepseekmath}, proposed by DeepSeek, is a \emph{critic-free} reinforcement learning algorithm. Given a query $q$, GRPO samples $G$ candidate outputs $\{o_1, \dots, o_G\}$ from the old policy $\pi_{\theta_{\text{old}}}$, and defines the normalized advantage function using the corresponding rewards $\{r_1, \dots, r_G\}$:
\begin{equation}
A_i = \frac{r_i - \text{mean}(\{r_1, r_2, \dots, r_G\})}{\text{std}(\{r_1, r_2, \dots, r_G\})}
\end{equation}

The policy $\pi_\theta$ is then updated by maximizing the following objective: \looseness=-1
% \begin{equation}
% J_{\text{GRPO}}(\theta) = 
% \mathbb{E}_{q, \{o_i\}} \left[ \frac{1}{G} \sum_{i=1}^G  \min \!\big( r_i(\theta) A_i,\; \text{clip}(r_i(\theta), 1-\epsilon, 1+\epsilon) A_i \big) \right] - \beta D_{\mathrm{KL}}(\pi_\theta \,\|\, \pi_{\text{ref}})
% \end{equation}
\begin{multline}
    J_{\text{GRPO}}(\theta) = \mathbb{E}_{q, \{o_i\}} \Bigg[ \frac{1}{G} \sum_{i=1}^G \min \Big( r_i(\theta) A_i, \\
    \text{clip}(r_i(\theta), 1-\epsilon, 1+\epsilon) A_i \Big) \Bigg] - \beta D_{\mathrm{KL}}(\pi_\theta \,\|\, \pi_{\text{ref}})
\end{multline}
where $\epsilon$ and $\beta$ are hyperparameters,  $r_i(\theta) = \tfrac{\pi_\theta(o_i \mid q)}{\pi_{\theta_{\text{old}}}(o_i \mid q)}$ is the importance sampling ratio, and 
% $D_{\mathrm{KL}}(\pi_\theta \,\|\, \pi_{\text{ref}}) = \tfrac{\pi_{\text{ref}}(o_i \mid q)}{\pi_\theta(o_i \mid q)} - \log \tfrac{\pi_{\text{ref}}(o_i \mid q)}{\pi_\theta(o_i \mid q)} - 1$ 
$D_{\mathrm{KL}}(\pi_\theta \,\|\, \pi_{\text{ref}})$ is the KL divergence regularization with respect to a reference model.

\subsection{Reinforcement Learning with Verifiable Rewards}
Reinforcement Learning with Verifiable Rewards (RLVR) is a framework that trains models using verifiable reward functions without relying on human feedback~\citep{lambert2024tulu}. In RLVR, the reward function is typically defined by deterministic rules that automatically assess the correctness of model outputs, providing binary signals (1 for correct, 0 for incorrect): \looseness=-1
\begin{equation}
r(y) = \texttt{verify}(y),
\end{equation}
where $\texttt{verify}(\cdot)$ is a verifiable function determining whether the output $y$ is correct.  

Depending on the verification source, rewards can be obtained in different ways. When ground truth labels $y^\star$ are available, the accuracy is determined by direct comparison $r(y) = \mathbf{1}[y = y^\star]$, as in Group Relative Policy Optimization (GRPO), where rules-based rewards check both the accuracy of the solutions and the required output format. In the absence of labels, verification can be performed in an unsupervised manner using self-consistency~\citep{wang2023selfconsistency, zuo2025ttrl}, where the majority-voted answer from a set of candidate outputs is treated as the correct answer and rewards are assigned accordingly. This formulation highlights that verifiable rewards can be constructed either with or without supervision, enabling reinforcement learning to be applied even in data-scarce or fully unsupervised reasoning scenarios.

%% file: sections/framework.tex
\section{Framework}
\label{sec:framework}

% \subsection{Overview}
In this section, we present a comprehensive overview of \framework, a collaborative Coach–Player paradigm for data-free reinforcement learning. \framework introduces a symbiotic learning loop between two independent language models: the \textbf{Coach}, a curriculum designer, and the \textbf{Player}, a reasoning solver.\looseness=-1

The core objective is to maximize learning progress without human-curated data. To achieve this, the Coach generates mathematical tasks tailored to the Player’s current capability, while the Player attempts to solve them. The key innovation lies in the cooperative reward mechanism: the Coach is optimized not to stump the Player, but to maximize the Player’s capability based on Coach-proposed tasks. This ensures that the curriculum remains instructive, learnable, and adaptive. \looseness=-1

We illustrate the main framework in \cref{fig:layer-architecture}, and the pseudo-code of algorithm can be found in ~\cref{sec:pseudo_code}. Formally, let $\pi_\theta^{\text{C}}$ denote the Coach policy and $\pi_\phi^{\text{P}}$ the Player policy. At each round $t$: \looseness=-1

\begin{enumerate}[leftmargin=10pt]
    \item \textbf{Coach designs plan.} 
    The Coach generates a batch of $m$ task instructions $\{x_i\}_{i=1}^m \sim \pi_{\theta_t}^{\text{C}}(\cdot)$, where $\pi_{\theta_t}^{\text{C}}$ is the current Coach policy.\looseness=-1
    
\input{fig_code/main}
    \item \textbf{Player executes training.}  
    For every \(x_i\) the current Player produces \(n\) independent answers  
    \(\{y_{i,j}\}_{j=1}^n \sim \pi^{\text{P}}_{\phi_t}(\cdot|x_i)\).  
    Majority voting over the \(n\) answers yields a \emph{pseudo-label} \( y_i^*\). Then each answer receives a verifiable reward $r_{i,j}= \mathbb{I}[y_{i,j}=y_i^*]$ 
    as well as a GRPO advantage \(A_{i,j}\) computed w.r.t. the \(n\) samples for question \(i\).  
    The \emph{instruction-level training reward} is obtained by averaging: \looseness=-1
    $R_i^{\text{Player}}=\frac{1}{n}\sum_{j=1}^n r_{i,j}$. The set \(\{(x_i,\{y_{i,j}\}_{j=1}^n)\}_{i=1}^m\) constitutes one GRPO batch, and Player parameters \(\phi_t\) are updated using GRPO method while keeping KL within a trust-region. \looseness=-1
    
    % \item \textbf{Player evaluates results.}  
    % The updated Player is evaluated on a fixed held-out validation set \(D_{\text{val}}\), yielding a \emph{progress reward} $\Delta_t = \text{Acc}_{\text{val}}(\pi_{\phi_{t+1}}^{\text{P}}; \mathcal{D}_{\text{val}}) - \text{Acc}_{\text{val}}(\pi_{\phi_t}^{\text{P}}; \mathcal{D}_{\text{val}})$, which measures the Player’s accuracy difference on the validation set \(\mathcal{D}_{\text{val}}\).

    \item \textbf{Player evaluates results.}  
    The updated Player receives \emph{environment feedback} (computed using a fixed evaluation set \( \mathcal{D}_{\text{val}} \)), yielding a \emph{progress reward}
    \[
        \Delta_t
        = \text{Acc}\!\left(\pi_{\phi_{t+1}}^{\text{P}}; \mathcal{D}_{\text{val}}\right)
        - \text{Acc}\!\left(\pi_{\phi_t}^{\text{P}}; \mathcal{D}_{\text{val}}\right),
    \]
    which measures the Player’s accuracy difference after receiving environment feedback.

    \item \textbf{Coach adjusts plan.}  
    Each instruction \(x_i\) is assigned an \emph{instruction reward} $R_i^{\text{Coach}} = R_i^{\text{Player}} \cdot \Delta_t$,
    i.e., instructions that produced high Player rewards and coincided with a global accuracy improvement are reinforced.  
    A group of $m$ instruction-level REINFORCE steps update Coach parameters \(\theta_t\) using each instance in the batch \(\{(x_i,R_i^{\text{Coach}})\}_{i=1}^m\).
\end{enumerate}

The entire loop is trained end-to-end with separate policy optimization for Coach and Player using the REINFORCE and GRPO, respectively. Critically, no human prompts and no external curricula are ever used. The Coach learns to teach, and the Player learns to solve, purely through interaction with each other. This cooperative design sidesteps the instability of adversarial self-play while retaining the benefits of open-ended, adaptive curriculum generation. In the following subsections, we detail the architecture, reward design, and training procedure of both the Coach and the Player.

\subsection{Coach} 
\label{subsec:Coach}

% The Coach serves as an adaptive \emph{curriculum designer}, fundamentally responsible for generating tasks that improve the Player's current reasoning capabilities. Unlike traditional static curriculum approaches, our Coach operates as a dynamic learning policy that continuously refines its task generation strategy based on the Player's learning trajectory. The Coach never sees ground-truth solutions, while its only feedback sign is the scalar $\Delta_t$, the process on the held-out validation set after the Player has been updated. 
The Coach serves as an adaptive \emph{curriculum designer}, fundamentally responsible for generating tasks that improve the Player's current reasoning capabilities. Unlike traditional static curriculum approaches, our Coach acts as a dynamic learning policy that continually refines its task-generation strategy in response to the Player's learning trajectory. \textbf{The Coach never observes ground-truth solutions}; instead, it receives only a scalar \emph{environment feedback} signal, $\Delta_t$, which captures the post-update performance change (computed using a fixed evaluation set $\mathcal{D}_{\text{val}}$). \looseness=-1

\paragraph{Difficulty-Filtered Batching}  
To ensure that every proposal task is \emph{learnable yet non-trivial}, we use a lightweight difficulty check during the task-generation phase. For each candidate task \(x_i\) sampled from \(\pi^{\text{C}}_\theta\), we rollout \(n\) Player answers $\{y_{i,j}\}_{j=1}^n \sim \pi^{\text{P}}_{\phi}(\cdot|x_i)$, obtain the majority-voted pseudo-label \(y_i^*\), and compute the rollout-dependent accuracy score of the instruction.

\begin{equation}
    \label{difficulty-filtered batching}
    acc_i = \frac{1}{n}\sum\nolimits_{j=1}^n \mathbb{I}[y_{i,j}=y_i^*].
\end{equation}

This score effectively measures the problem's alignment with the Player's current capability frontier. The Coach then applies a principled filtering criterion, retaining only problems whose accuracy scores fall within the pedagogically optimal zone of $0.2\leq acc_i\leq 0.8$. Problems outside this range are immediately discarded and replaced through on-the-fly resampling. This online filter guaranties that the final mini-batch of \(m\) questions is challenging enough to promote skill development yet solvable enough to avoid frustration, providing a natural curriculum ramp.

\paragraph{Design Objectives}  
The Coach embodies a learner-centered educational philosophy, where its primary objective is to optimize the constructiveness of the proposed-task for the Player. 
% The Coach's training objective is formulated to directly optimize for the Player's learning progress. 
Formally, the Coach policy $\pi_\theta^{\text{C}}$ is optimized using instruction-level rewards that combine local training effectiveness with global educational outcomes. 
\begin{equation}
    R_i^{\text{Coach}} = R_i^{\text{Player}} \cdot \Delta_t
\end{equation}
where 
\begin{equation}
R_i^{\text{Player}} = \frac{1}{n}\sum_{j=1}^n r_{i,j}
\end{equation}
represents the average training reward achieved by the Player on instruction $x_i$, and 
% \begin{equation}
% \Delta_t = \text{Acc}_{\text{val}}(\pi_{\phi_{t+1}}^{\text{P}}) - \text{Acc}_{\text{val}}(\pi_{\phi_t}^{\text{P}})
% \end{equation}
$\Delta_t = \text{Acc}_{\text{val}}(\pi_{\phi_{t+1}}^{\text{P}}) - \text{Acc}_{\text{val}}(\pi_{\phi_t}^{\text{P}})$
measures the Player’s accuracy improvement after receiving environment feedback

This multiplicative reward embodies a pedagogical principle: proposed tasks receive positive reinforcement only when they \textbf{simultaneously achieve high Player performance during training (high $R_i^{\text{Player}}$) and contribute to measurable learning progress (positive $\Delta_t$)}. The Coach parameters are updated through REINFORCE using the batch of instruction-reward pairs $\{(x_i, R_i^{\text{Coach}})\}_{i=1}^m$:
\begin{equation}
    \nabla_\theta J(\theta) = \frac{1}{m}\sum_{i=1}^m R_i^{\text{Coach}} \nabla_\theta \log \pi_\theta^{\text{C}}(x_i).
\end{equation}
% This objective function naturally aligns the Coach's incentives with effective pedagogy: the instructions must satisfy both immediate learnability and long-term educational value. Therefore, the Coach learns to create instructional experiences that systematically advance the Player's mathematical reasoning capabilities.
% First, the multiplicative structure ensures that problems must satisfy both immediate learnability and long-term educational value. 
%Second, the global progress signal $\Delta_t$ provides a quality filter that distinguishes between superficial task completion and genuine skill acquisition. Third, the instruction-level granularity enables the Coach to learn nuanced associations between specific problem characteristics and their educational effectiveness. 
% Importantly, this formulation captures the essence of adaptive tutoring: the Coach learns to create instructional experiences that systematically advance the Player's mathematical reasoning capabilities, naturally discouraging problems that are either too trivial or too difficult while reinforcing those that maintain optimal cognitive demand.

\subsection{Player} 
\label{subsec:Player}

The Player functions as the primary reasoning model, designed to develop robust mathematical problem-solving capabilities through iterative interaction with the Coach-generated curriculum. 

\paragraph{Design Objectives and Collaborative Dynamics}
The Player's core objective is to maximize solving accuracy on mathematical problems while developing generalizable reasoning strategies. The Player operates within a collaborative learning framework where its performance directly influences curriculum adaptation through a sophisticated feedback mechanism. The Player's learning process is also inherently adaptive, continuously calibrating its problem-solving strategies based on feedback from the Coach-generated curriculum.

The Player employs multi-sample reasoning for each problem $x_i$, generating $n$ independent solution attempts $\{y_{i,j}\}_{j=1}^n$ to enable robust pseudo-label generation through majority voting. This approach mitigates individual reasoning errors, provides confidence estimates for generated solutions, and creates multiple learning signals from each instructional instance.

The interaction protocol between the Coach and Player establishes a dynamic feedback loop that drives mutual improvement. This ensures the curriculum remains at an optimal difficulty, maintaining learning momentum and continuously pushing the frontier of the Player's capabilities.

\paragraph{Training and Optimization} 
The Player is optimized using GRPO, which enables stable learning from the pseudo-labels generated through majority voting. For each problem instance $x_i$, the Player receives rewards 
\begin{equation}
    r_{i,j} = \mathbb{I}[y_{i,j} = y_i^*],
\end{equation}
where $y_i^*$ is the majority-voted pseudo-label. 
The GRPO advantage computation considers the relative performance across the $n$ samples for each problem:
% and $\beta$ controls the KL divergence penalty relative to a reference policy $\pi_{\text{ref}}$. 

\begin{equation}
A_{i,j} = \frac{r_{i,j} - \text{mean}(\{r_{i,1}, r_{i,2}, \dots, r_{i,n}\})}{\text{std}(\{r_{i,1}, r_{i,2}, \dots, r_{i,n}\})}
\end{equation}

% This formulation encourages the Player to strengthen its most promising reasoning paths while maintaining diversity in problem-solving approaches, fostering both accuracy and robustness in mathematical reasoning.
% The optimization process incorporates curriculum-aware learning that prioritizes challenging yet solvable problems, ensuring that skill development remains aligned with the Coach's instructional design and creating synergistic learning dynamics that benefit both agents.

% Together, the two subsections complete the \framework training loop: the Coach proposes, the Player explores, and the scalar improvement signal closes the cooperative curriculum-design game. Through this carefully orchestrated Coach-Player interaction, the system achieves data-free mathematical reasoning development, where both agents co-evolve to maximize learning efficiency without requiring human-curated training data or explicit curriculum design.
This collaborative process completes the \framework training loop: the Coach designs training curriculum, the Player explores potential solutions, and the Player's consequent capability guides the curriculum's evolution. 
% This formulation encourages the Player to reinforce its most successful reasoning paths while maintaining exploration, fostering both accuracy and robustness. 
The process is inherently curriculum-aware, prioritizing challenging yet solvable problems to ensure the Player’s skill development remains aligned with the Coach’s adaptive strategy. Through this orchestrated interaction, the framework achieves data-free mathematical reasoning development, where both models co-evolve to maximize learning efficiency without reliance on human-curated data or a pre-defined curriculum.

%% file: fig_code/main.tex
% \begin{figure}
%     \centering
%     \includegraphics[width=\textwidth]{figure/main_version_6.png}
%     \caption{The illustration on the conceptual layered architecture on the design of \framework. The iterative process includes four stages. \textbf{Coach designs plan}: The coach gives instructions of suitable difficulty based on the player’s current ability. \textbf{Player executes training}: The player executes each instruction multiple times, uses majority voting to get pseudo-labels, and updates with GRPO. \textbf{Player evaluates results}: The updated player is tested on a prepared validation set, and the accuracy of validation is recorded. \textbf{Coach adjusts plan}: The coach updates with REINFORCE, using the player’s performance on both the proposed instructions and the validation set as rewards.}
%     \label{fig:layer-architecture}
%     \vspace{-1em}
% \end{figure}

\begin{figure*}[t]  % 关键修改：加了星号 *，[t] 表示优先顶端对齐
    \centering
    % width=\textwidth 在 figure* 环境下表示整个页面的宽度（双栏宽度+间距）
    \includegraphics[width=0.8\textwidth]{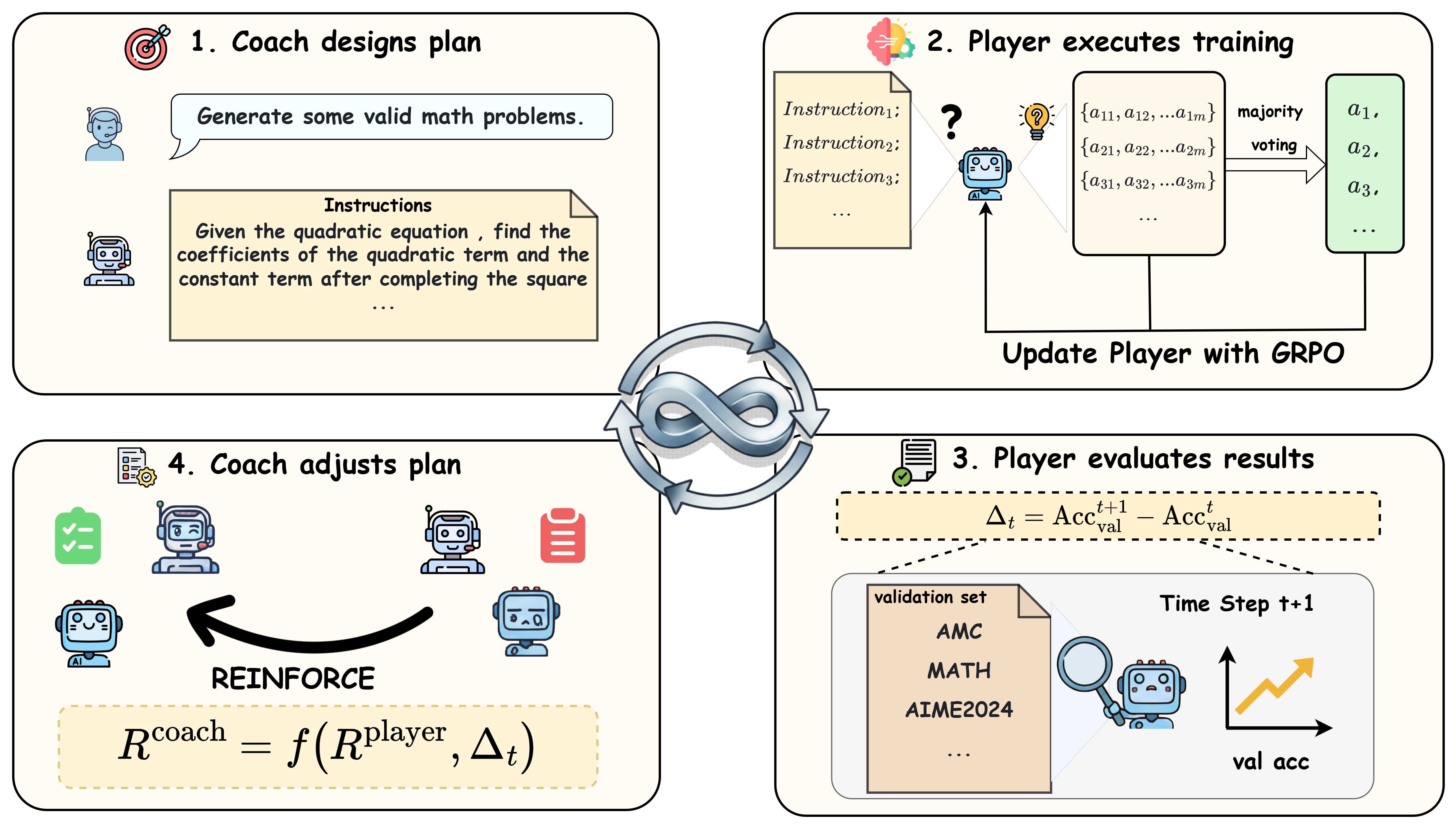}
    
    \caption{The illustration on the conceptual layered architecture on the design of \framework. The iterative process includes four stages. \textbf{Coach designs plan}: The coach gives instructions of suitable difficulty based on the player’s current ability. \textbf{Player executes training}: The player executes each instruction multiple times, uses majority voting to get pseudo-labels, and updates with GRPO. \textbf{Player evaluates results}: The updated player is interacted on a prepared environment, and the environment feedback-induced accuracy gap is recorded. \textbf{Coach adjusts plan}: The coach updates with REINFORCE, using the player’s performance on both the proposed instructions and the environment feedback as rewards.}
    
    \label{fig:layer-architecture}
    % \vspace{-1em}
\end{figure*}  % 别忘了结束也要加星号

%% file: sections/experiment.tex
\section{Experiments}
\label{sec:experiments}

\subsection{Experiment Setup}
\label{sec:experiment_setup}
\textbf{Coach Model Selection.} We fix the Coach to \textbf{Qwen2.5-Math-7B-Instruct ~\citep{yang2024qwen2} that is further warmed up with 4K PRIME Eurus-2-RL-Data~\citep{cui2025process}}.
This warm-up is mainly used to strengthen the Coach's ability to pose constructive, diagnostically useful questions, which in turn improves the quality of the curriculum it generates and facilitates effective co-evolution with the Player.
Importantly, our use of warm-up data does not contradict the ``no external training data'' setting in the co-evolution stage. No additional external training data is introduced \emph{after} the warm-up, and all subsequent learning signals arise from the Coach and Player interaction and the environment feedback. Our setting is \textbf{\emph{data-free Player training with an optionally warmed-up Coach}}.
\input{table/model_train}

\textbf{Player Model Selection.}  We select four base models for our training experiments, representing the three main stages of a typical LLM training lifecycle: pre-training, supervised fine-tuning (SFT), and reinforcement learning.
% First, we seek foundation models that have undergone extensive math-specialized pre-training. Second, we consider the depth of mathematical knowledge acquired through extensive SFT procedures. Third, we include models that have undergone reinforcement learning training to capture different optimization paradigms. 
% The direct comparisons are:
\begin{itemize}[itemsep=2pt, parsep=0pt, topsep=0pt, partopsep=0pt]
    \item \textbf{Qwen2.5-Math-1.5B}~\citep{yang2024qwen2}: a mathematical pre-training model.
    \item \textbf{OpenMath-Nemotron-1.5B}~\citep{moshkov2025aimo2}: a large-scale SFT enhanced model based on Qwen2.5-Math-1.5B.
    \item \textbf{Qwen2.5-Math-7B-Instruct}~\citep{yang2024qwen2} and \textbf{OctoThinker-3B-Hybrid-Zero}~\citep{wang2025octothinker}: models optimized through reinforcement learning.
\end{itemize}
More details about these models are introduced in \cref{appendix:model_selection}.

\textbf{Training Details.} All experiments were conducted within the verl~\citep{sheng2025hybridflow}. We use AMC as the fixed held-out validation $D_{val}$ during training. We choose AMC because its difficulty is typically moderate, it is neither trivial which would quickly saturate and yield a near-zero learning signal nor overly hard which would produce sparse, noisy feedback early on. As a result, AMC provides a more stable and informative progress signal for computing the environment feedback $\Delta_t$ throughout training. All experiments were conducted using 4 to 8 NVIDIA A800-80GB GPUs per setting. We set the batch size as 16 and the number of rollout samples for each prompt as 16, ensuring that each training round involves the Coach generating 16 questions and the Player producing 16 candidate solutions for majority voting-based pseudo-label generation. More hyperparameter configurations and prompt templates are provided in ~\cref{appendix:training_details}.\looseness=-1

\textbf{Evaluation Details.} We evaluate the Player models on six established mathematical reasoning benchmarks spanning diverse difficulty levels: AMC, Minerva~\citep{Lewkowycz2022SolvingQR}, MATH-500~\citep{hendrycks2021measuring}, Olympiad-Bench~\citep{he-etal-2024-olympiadbench}, and AIME 2024 and AIME 2025. To ensure robustness, we employ benchmark-specific sampling strategies calibrated to each benchmark's difficulty: mean@32 for AIME benchmarks, mean@10 for AMC, mean@6 for Minerva, mean@5 for MATH-500, and mean@3 for Olympiad-Bench. Since AMC is used as the validation set during training, we compute both the average score on all six datasets and the OOD average score on the other five datasets except for AMC. All sampling settings are kept consistent with the training configuration, as illustrated in ~\cref{appendix:training_details}. 
% The validation sets from these benchmarks also could serve as evaluation data for monitoring Player model progress during training, ensuring alignment between training objectives and final evaluation metrics. \looseness=-1

\textbf{Baselines.} For our main experiments, beyond the selected base models, we considered two representative unsupervised training paradigms as baselines. The first is RENT~\citep{prabhudesai2025maximizing}, which employs entropy minimization: the model’s own confidence in its generated answers is treated as a reward signal, without relying on external feedback. The second is R-Zero~\citep{huang2025rzeroselfevolvingreasoningllm}, which initializes two roles of the same model that interact adversarially, with the challenger generating tasks and the solver attempting to solve them. 
% All baseline methods' training data and initialization settings are summarized in \cref{sec:appendix}.

\input{fig_code/critic-score-mean}
\subsection{Results}
% \subsubsection{Baseline}
% \subsubsection{Results in out-of-distribution Mathematical Reasoning}
We present the main results in Table \ref{tab:main}. We have the following observations:\looseness=-1

\paragraph{\framework outperforms other unsupervised RL methods:}
The results show that \framework achieves superior performance compared to other unsupervised RL baselines, consistently achieving the highest overall average and OOD average scores across all four base model. Impressively, \framework successfully improves high-performing base models OpenMath-Nemotron-1.5B (from $59.5$ to $62.1$). Notably, we found that the method from R-Zero failed on OpenMath-Nemotron-1.5B, failing to be trained as a Challenger as required by R-Zero. This demonstrates \framework's ability to push models beyond their apparent performance ceiling, a critical advantage for practical applications where starting from pre-optimized models is common. \looseness=-1

\paragraph{Strong out-of-distribution generalization:}
% Since \framework and RENT use signals from the AMC dataset during training, all other benchmarks represent out-of-distribution evaluation scenarios. 
\framework achieves better OOD average scores across all four tested models, demonstrating that the reasoning capabilities learned from AMC competition problems effectively transfer to diverse mathematical domains. On MATH, \framework consistently outperforms other methods with improvements ranging from $1.8$ to $6.9$ points over base models. The most striking OOD generalization occurs on the Minerva benchmark, where \framework achieves obvious improvements: from $16.3$ to $28.0$ ($71.8\%$) on Qwen2.5-Math-1.5B and $34.6$ to $44.9$ ($29.8\%$) on Qwen2.5-Math-7B-Instruct. \looseness=-1

\paragraph{Performance analysis for different initial models:}
The experimental results reveal distinct performance patterns that correlate with initial model characteristics. (1) Foundation models demonstrate high improvement potential: Qwen2.5-Math-1.5B achieves an overall $5.5$ points improvement ($23.6\%$ relative gain), suggesting that models with domain-specific pre-training provide strong foundations for \framework's optimization approach. (2) SFT-enhanced models show diminishing but meaningful returns: Despite starting from a high $59.5$ points baseline after extensive SFT on 5.5 million instances, OpenMath-Nemotron-1.5B still achieves an overall  $2.6$ points improvement, demonstrating \framework's ability to push beyond traditional SFT limits. (3) RL-optimized models exhibit varied enhancement: Qwen2.5-Math-7B-Instruct shows remarkable $4.9$ points improvement despite instruction tuning, while OctoThinker-3B-Hybrid-Zero shows modest $2.3$ points gains. 

\paragraph{Robustness to different held-out environments.} To examine whether the effectiveness of \framework depends on the choice of the held-out validation environment, we further replace AMC with Minerva and OlympiadBench as the validation environment for computing the Coach reward signal. As shown in Appendix~\ref{appendix:new_validation}, \framework still brings consistent improvements on both Qwen2.5-Math-1.5B and Qwen2.5-Math-7B-Instruct. These results suggest that the proposed Coach--Player paradigm is not merely exploiting dataset-specific patterns from AMC, but can generalize across different validation environments and continue to improve the Player's reasoning ability.
% \paragraph{Patterns observed during \framework training}
% A critical pattern observed in the \framework training was the emergence of an autonomous learning curriculum, where the Coach model progressively generated questions of increasing difficulty. This dynamic arose naturally from the Coach's objective to challenge the Player, causing it to move beyond the Player's mastered skills and propose more complex problems as training advanced. This pattern of escalating difficulty proved highly beneficial, as it consistently kept the Player operating within its optimal learning zone. By smoothly guiding the Player from simple to complex challenges, this self-adjusting mechanism promoted robust skill acquisition and was a key factor in the stable and efficient training of the Player.

% In this section, we perform comprehensive analyses to examine the behavior and effectiveness of our \framework framework. For consistency across all analytical experiments, we employ the Qwen2.5-MATH-1.5B model as our base architecture.
\input{table/ablation}
\input{fig_code/train_dynamic}
\subsection{Training Dynamics}
We analyze the training dynamics of \framework by tracking both validation accuracy on AMC and the consistency of the Player’s responses throughout training steps. As shown in Fig~\ref{fig:train_dynamic}, \framework steadily improves the Player’s performance across all four base models, indicating that the cooperative Coach–Player optimization loop enables stable and continual reasoning enhancement. The performance gains are gradual yet consistent, demonstrating that the curriculum adapts effectively to the Player’s evolving capabilities. \looseness=-1

Fig~\ref{fig:critic_score} illustrates the evolution of answer consistency, where lower values correspond to more challenging tasks proposed by the Coach. Notably, for Qwen2.5-Math-1.5B and OpenMath-Memotron-1.5B, two reasoning models without previous RL training, the downward trends in consistency indicates that the Coach progressively generated questions of increasing difficulty, maintaining the Player within an optimal learning zone. For OctoThinker-3B-Hybrid-Zero and Qwen2.5-Math-7B-Instruct, with better performance benefiting from previous RL training, the difficulty maintains a reasonably range. \looseness=-1

Additionally, we found that the length of problems proposed by the Coach is increasing, indicating that the Coach gradually generates more complex tasks to adapt to the Player's growing capabilities. Meanwhile, the Player's response length is decreasing, suggesting that the Player is generating increasingly efficient answers.
Details can be found in ~\cref{appendix:length} and ~\cref{appendix:problems}. Together, these results highlight that \framework not only drives performance improvement but also naturally induces a self-adjusting curriculum based on the Player's performance.\looseness=-1

\subsection{Ablation Study}
\input{fig_code/ablation_dynamic}

To systematically evaluate the individual contributions of each core component within \framework, we conduct a thorough ablation study on the Qwen2.5-MATH-1.5B model. We examine the relative importance of three critical modules (i.e., Coach update, Coach SFT warm-up, and instruction filter) by selectively removing each component and measuring the resulting performance degradation across multiple mathematical reasoning benchmarks. The comprehensive results of this ablation analysis are presented in Table \ref{tab:ablation}, and training dynamics of different ablations are shown in Fig \ref{fig:ablation_dynamic}.\looseness=-1

\textbf{Coach Update.} ablation fixes the Coach model throughout training instead of adapting it to the Player's evolving performance. This dynamic adaptation mechanism enables personalized curriculum generation tailored to the Player's current capabilities, creating a co-evolutionary learning dynamic. Removing Coach updates degrades average accuracy from 28.8\% to 25.3\%, with out-of-distribution (OOD) performance dropping from 26.8\% to 23.1\%, demonstrating the critical importance of adaptive instruction.

\textbf{Coach Warm-up.} ablation eliminates the initial warm-up phase. This initialization ensures that the Coach can generate high-quality math problems from the outset, establishing a strong foundation for subsequent cooperative training. Without warm-up, average accuracy drops to 23.7\% (OOD: 21.2\%), indicating that proper Coach initialization is essential for effective curriculum generation.

\textbf{Instruction Filter.} ablation removes the difficulty calibration mechanism that maintains problems within the optimal learning zone, where the accuracy is between 0.2 and 0.8. This filter ensures generated problems remain challenging yet solvable, maintaining the Player at its capability frontier. Disabling this mechanism reduces average accuracy to 24.9\% (OOD: 22.5\%), confirming that appropriate difficulty calibration is crucial for efficient learning.

% These results demonstrate that each component contributes uniquely to \framework's effectiveness, with their synergistic interaction enabling superior mathematical reasoning development compared to any subset of the framework.

%% file: table/model_train.tex
\begin{table*}[t!]
    \centering
    \caption{Performance comparison between \framework and baseline methods on mathematical reasoning benchmarks.
    % We evaluate our method against RENT and R-Zero baselines. 
    Overall Average indicates the mean performance over all benchmarks. OOD Average refers to the out-of-distribution performance, computed as the mean across all benchmarks except the AMC datasets, because RENT was trained on AMC and \framework validation also used AMC. This separation enables a fair comparison by clearly distinguishing in-distribution (AMC) results from out-of-distribution generalization performance. Bold values indicate best performance for each metric.}
    \resizebox{\textwidth}{!}{
    \begin{tabular}{l >{\columncolor{blue!9}}c >{\columncolor{blue!9}}c >{\columncolor{yellow!9}}c c c c c c}
        \toprule
        \textbf{Models} & \textbf{Average} &\textbf{OOD Average} & \textbf{AMC} & \textbf{AIME 2024} & \textbf{AIME 2025} &  \textbf{Minerva} & \textbf{MATH} & \textbf{Olympiad}\\
        % Human & 93.90 & 91.80 & 87.30 \\
        \midrule
        \textit{Qwen2.5-Math-1.5B} \\
        \quad Base Model & 23.3 & 19.8 & 34.6 & 6.2 & 2.8 & 16.3 & 56.2 & 23.4\\
        \quad R-Zero (Iter 3) & 27.1 & 24.7 & 39.2 & 9.8 & 5.0  & 19.3 & 62.4 & 26.8\\
        \quad RENT &  27.1 & 24.7& 39.3 & \textbf{10.0} & 5.0  & 19.0 & 62.2 & \textbf{27.1}\\
        \quad \textit{\textbf{\framework}} & \textbf{28.8}& \textbf{26.8}& \textbf{39.4} & 9.8 & \textbf{5.4} & \textbf{28.0} & \textbf{63.1} & 26.9\\
        
        \midrule
        \textit{OpenMath-Nemotron-1.5B} \\
        \quad Base Model & 59.5 & 54.9& 82.3 & \textbf{55.6} & 43.3 & \textbf{25.1} & 89.4 & 61.0\\
        \quad R-Zero (Iter 3) & - & - & - & - & - & - & - & -\\
        \quad RENT & 61.7 & 56.5& \textbf{87.7} & 55.0 & 46.0  & 24.2 & 90.7 & 66.7\\
        \quad \textit{\textbf{\framework}} & \textbf{62.1} & \textbf{57.0}& 87.5  & 54.9 & \textbf{46.9} & 24.3 & \textbf{91.2} & \textbf{67.9} \\ 
        \midrule
        \textit{OctoThinker-3B-Hybrid-Zero} \\
        \quad Base Model & 21.3 & 20.6& 24.6  &  3.9 & 1.7  & 16.3 & 57.9 & 23.4 \\
        \quad R-Zero (Iter 3) & 20.5 & 19.5& 25.9 & 2.0 & 0.3 & 14.6 & 58.1 & 22.3 \\
        \quad RENT & 23.0 & 21.7 & \textbf{29.2} & \textbf{7.3} & \textbf{2.1} & 15.0 & 60.2 & 24.1\\
        \quad \textit{\textbf{\framework}} & \textbf{23.6} & \textbf{22.0} & 28.0  & 4.8 & 1.7 & \textbf{22.1} & \textbf{60.4} & \textbf{24.7}\\
        % Huggingface-Agents & 30.19 & 11.63 & 7.69 & 16.97\\ \\
        \midrule
        \textit{Qwen2.5-Math-7B-Instruct} \\
        \quad Base Model & 35.8 & 33.0& 49.2 & 9.0 & 6.3 & 34.6 & 78.0 & 37.4 \\
        \quad R-Zero (Iter 3) & 36.9 & 34.2& 50.5 & 9.5 & 7.4 & 32.7 & 83.3 & 38.1\\
        \quad RENT & 39.2 & 37.6& 53.1 & 10.8 & \textbf{9.9} & 38.8 & 83.8 & \textbf{38.8}\\
        \quad \textit{\textbf{\framework}} & \textbf{40.7} & \textbf{38.4} & \textbf{55.6} & \textbf{11.8}  & 9.6 & \textbf{44.9} & \textbf{84.2} & 38.3\\
        \bottomrule
    \end{tabular}}

    \label{tab:main}
\end{table*}

%% file: fig_code/critic-score-mean.tex
\begin{figure}[h]
    \centering
    \includegraphics[width=0.5\textwidth]{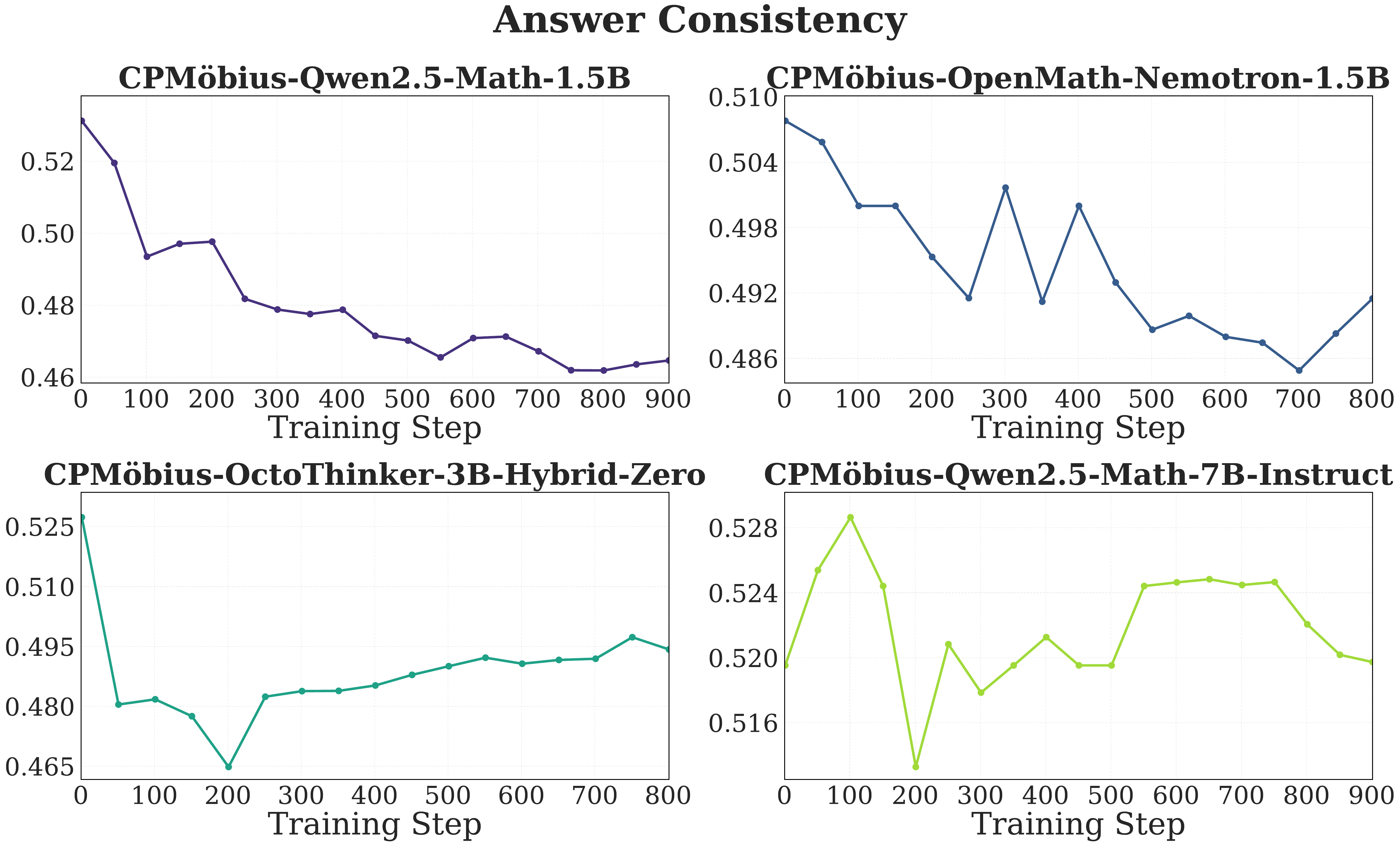}
    \caption{Visualization of the Player's answer consistency on Coach proposed tasks during training. A lower value indicates higher difficulty of the instructions.}
    \label{fig:critic_score}
    % \vspace{-1em}
\end{figure}

%% file: table/ablation.tex
\begin{table*}[t]
    \centering
    \caption{Ablation study results are based on the Qwen2.5-Math-1.5B base model. \textbf{w/o Coach Update}: disables training of the Coach. \textbf{w/o Coach Warm-up}: uses the base model as the Coach. \textbf{w/o Instruction Filter}: disables difficulty filtering by the Coach.}
    \resizebox{\textwidth}{!}{
    \begin{tabular}{l >{\columncolor{black!9}}c >{\columncolor{black!9}}c c c c c c c}
        \toprule
        \textbf{Models} & \textbf{Average} &\textbf{OOD Average} & \textbf{AMC} & \textbf{AIME 2024} & \textbf{AIME 2025} &  \textbf{Minerva} & \textbf{MATH} & \textbf{Olympiad}\\
        % Human & 93.90 & 91.80 & 87.30 \\
        \midrule
        \textit{Qwen2.5-Math-1.5B} \\
        \quad Base Model & 23.3 & 19.8 & 34.6 & 6.2 & 2.8 & 16.3 & 56.2 & 23.4 \\
        \quad \textit{\textbf{\framework}} & \textbf{28.8}& \textbf{26.8}& \textbf{39.4} & \textbf{9.8} & \textbf{5.4} & \textbf{28.0} & \textbf{63.1} & \textbf{26.9}\\
        \midrule
        \quad Ablation \\
        % \quad  $\vdash$~w/o Player Policy Reset & 25.0 & 10.0 & 5.1 & 41.7 & 15.8 & 65.3 & 28.6\\
        % \quad  $\vdash$~w/o Player Policy Reset &  & 10.3 & 5.7 & 40.5 & 15.4 & 64.3 & 27.9\\
        \quad $\vdash$~w/o Coach Update & 25.3 & 23.1 & 36.7& 8.7 & 4.8 & 17.2 & 58.4 & 26.3\\
        \quad $\vdash$~w/o Coach Warm-up & 23.7 & 21.2 & 36.1 & 9.2 & 3.6  & 13.8 & 54.4 & 24.8\\
        \quad $\vdash$~w/o Instruction Filter &  24.9 & 22.5 & 37.3 & 9.0 & 3.5 & 16.6 & 58.4 & 24.9\\
        \bottomrule
     \end{tabular}}
    \label{tab:ablation}
\end{table*}

%% file: fig_code/train_dynamic.tex
\begin{figure}[t]
    \centering
    \includegraphics[width=0.5\textwidth]{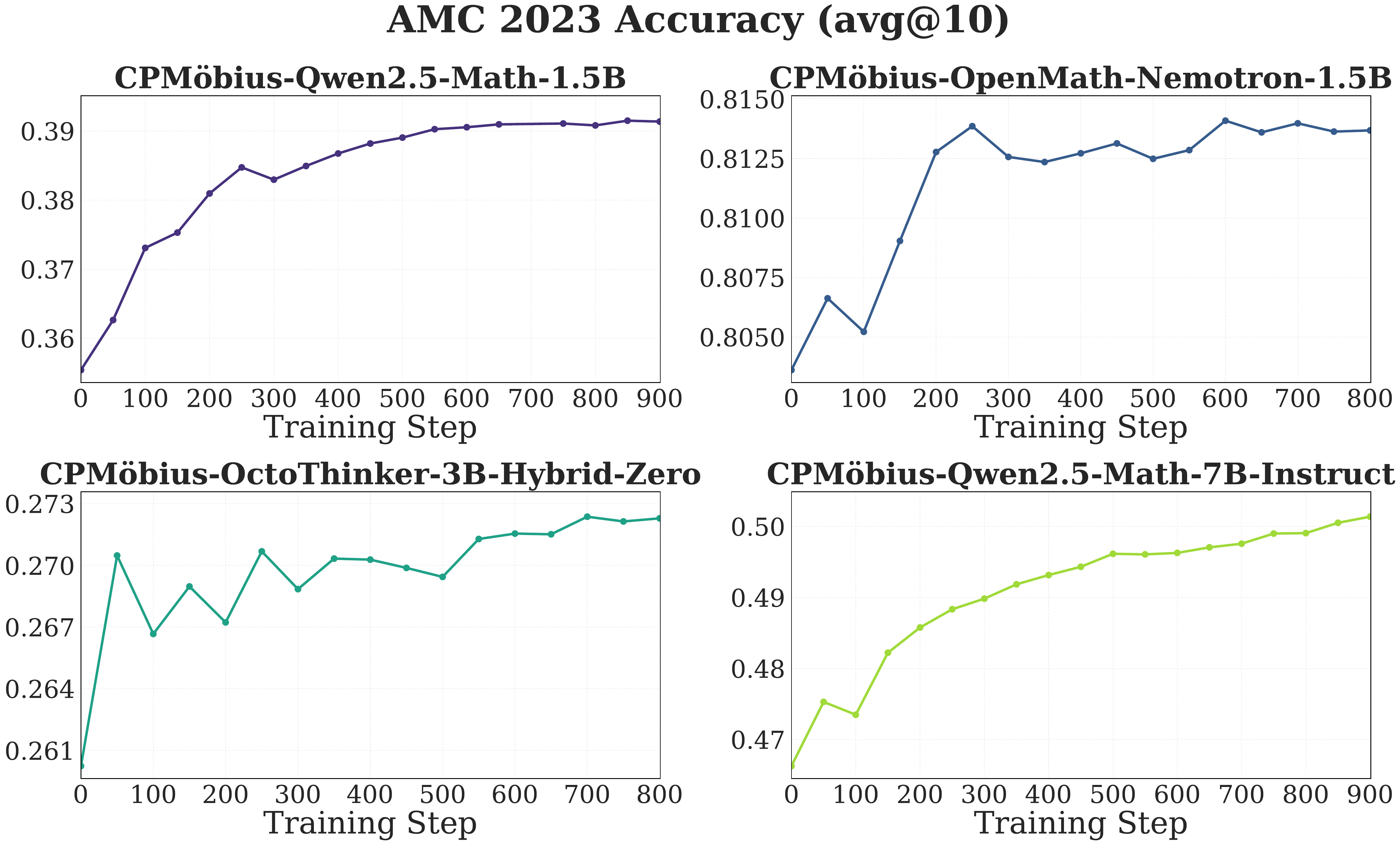}
    \caption{Visualization of the training dynamics of \framework using validation results on AMC dataset. The curves are smoothed with Time Weighted EMA, where \framework shows consistent performance improvement for different base models.}
    \label{fig:train_dynamic}
    % \vspace{-2.5em}
\end{figure}

%% file: fig_code/ablation_dynamic.tex
\begin{figure}[t]
    \centering
    \includegraphics[width=0.5\textwidth]{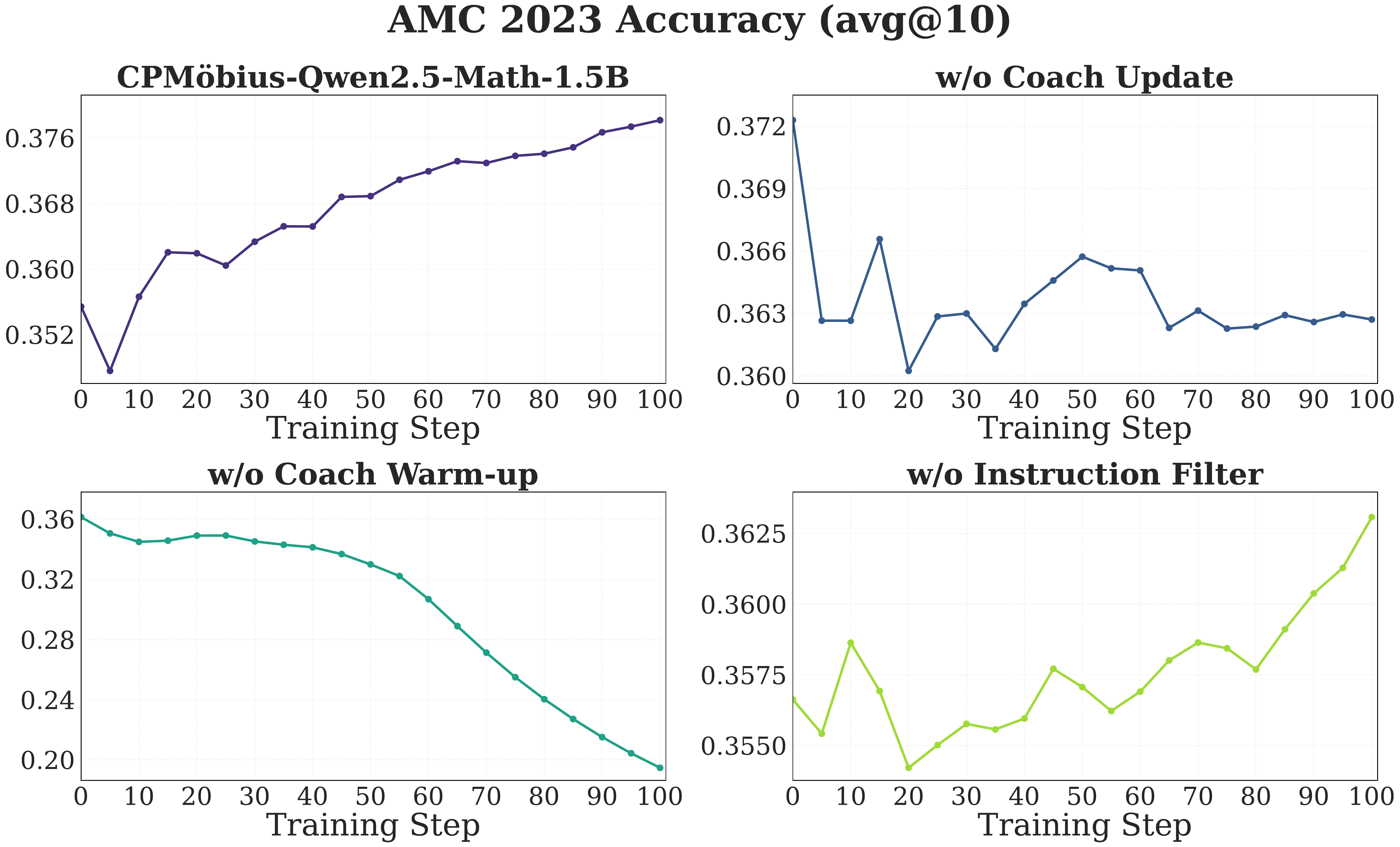}
    \caption{Visualization of the training dynamics on \framework and different ablation experiments using validation results on AMC dataset.}
    \label{fig:ablation_dynamic}
    % \vspace{-2.5em}
\end{figure}

%% file: sections/conclusion.tex
\section{Conclusion}
\label{sec:conclusion}

In this work, we introduced \framework, a novel Coach-Player framework inspired by multi-agent collaboration to enhance reasoning capabilities in a fully data-free setting. The central innovation of our framework is a collaborative optimization loop in which a coach model constructs a targeted curriculum that is rewarded based on the Player’s learning progress. This interaction naturally uncovers a curriculum that adapts to and evolves with the Player’s growing capabilities, successfully decoupling reasoning enhancement from without depending on previously defined tasks or human-curated labels. Our work demonstrates that a collaborative, data-free reinforcement learning strategy can be a powerful and efficient substitute training framework. 
% By showing that sophisticated reasoning skills can emerge from structured cooperation, we present a scalable and sustainable path forward for developing increasingly capable and autonomous AI systems. 
Future work could investigate applying the collaborative co-evolving paradigm to additional complex domains. Furthermore, examining the emergent behaviors and long-term stability of the interactions between co-evolving models represents a promising direction for future research.

%% file: sections/appendix.tex
\newpage
\section{Appendix}
\label{sec:appendix}
% \subsection{Use of LLM}
% In preparing this work, we use large language models (LLMs) to assist in writing and editing. Specifically, LLMs were employed to help refine the clarity and readability of certain sections and check for consistency in terminology. All technical content, experimental design, implementation details, and results were produced, verified, and analyzed independently by the authors. No part of the experimental process, including model training, data handling, or evaluation, relied on external LLM outputs.

\subsection{Pseudo-code for \framework}
\label{sec:pseudo_code}
\begin{algorithm}[H]
\caption{Coach-Player Framework for Data-Free Reinforcement Learning}
\label{alg:coachplayer}
\begin{algorithmic}[1]
\REQUIRE Pretrained Coach LLM $\pi_{\theta_0}^C$; Player LLM $\pi_{\phi_0}^P$; validation set $\mathcal{D}_{val}$
\REQUIRE Batch size $m$; samples per task $n$; iterations $T$; learning rates $\alpha_C, \alpha_P$
\STATE $\theta \leftarrow \theta_0, \phi \leftarrow \phi_0$ \hfill $\triangleright$ Initialize parameters
\FOR{$t \leftarrow 1$ \TO $T$}
    \STATE $\mathcal{B} \leftarrow \emptyset$ \hfill $\triangleright$ \textbf{COACH GENERATION PHASE}
    \WHILE{$|\mathcal{B}| < m$}
        \STATE $x_{cand} \sim \pi_{\theta}^C(\cdot)$ \hfill $\triangleright$ Coach proposes candidate task
        \STATE $\{y_j\}_{j=1}^n \sim \pi_{\phi}^P(\cdot|x_{cand})$ \hfill $\triangleright$ Player attempts task
        \STATE $y^* \leftarrow \text{MajorityVote}(\{y_j\}_{j=1}^n)$ \hfill $\triangleright$ Compute pseudo-label
        \STATE $acc \leftarrow \frac{1}{n}\sum_{j=1}^n \mathbb{I}[y_j = y^*]$ \hfill $\triangleright$ Calculate accuracy
        \IF{$0.2 \leq \text{acc} \leq 0.8$}
            \STATE $\mathcal{B} \leftarrow \mathcal{B} \cup \{x_{\text{cand}}\}$ \hfill $\triangleright$ Accept task if difficulty appropriate
        \ENDIF
    \ENDWHILE
    \hfill $\triangleright$ \textbf{PLAYER TRAINING PHASE}
    \FOR{$i \leftarrow 1$ \TO $m$}
        \STATE $\{y_{i,j}\}_{j=1}^n \sim \pi_{\phi}^P(\cdot|x_i)$ where $x_i \in \mathcal{B}$ \hfill $\triangleright$ Generate responses
        \STATE $y_i^* \leftarrow \text{MajorityVote}(\{y_{i,j}\}_{j=1}^n)$ \hfill $\triangleright$ Pseudo-label
        \STATE $r_{i,j} \leftarrow \mathbb{I}[y_{i,j} = y_i^*]$ for $j = 1, \ldots, n$ \hfill $\triangleright$ Assign rewards
        \STATE $A_{i,j} \leftarrow \frac{r_{i,j} - \bar{r}_i}{\sigma_i + \epsilon}$ \hfill $\triangleright$ GRPO advantages
        \STATE $R_i^{Player} \leftarrow \frac{1}{n}\sum_{j=1}^n r_{i,j}$ \hfill $\triangleright$ Instruction-level reward
    \ENDFOR
    \STATE $\phi \leftarrow \phi + \alpha_P \cdot \nabla_\phi \mathcal{L}_{GRPO}$ \hfill $\triangleright$ Update Player via GRPO
    \STATE $\Delta_t \leftarrow \text{Acc}_{val}(\pi_{\phi}^P; \mathcal{D}_{val}) - \text{Acc}_{val}(\pi_{\phi_{old}}^P; \mathcal{D}_{val})$ \hfill $\triangleright$ \textbf{EVALUATION PHASE}
    \FOR{$i \leftarrow 1$ \TO $m$} 
        \STATE $R_i^{Coach} \leftarrow R_i^{Player} \cdot \Delta_t$ \hfill $\triangleright$ Coach instruction reward
    \ENDFOR
    \hfill $\triangleright$ \textbf{COACH UPDATE PHASE}
    \STATE $\theta \leftarrow \theta + \alpha_C \cdot \frac{1}{m}\sum_{i=1}^m R_i^{Coach} \nabla_\theta \log \pi_{\theta}^C(x_i)$ \hfill $\triangleright$ REINFORCE update
\ENDFOR \\
\RETURN $\pi_{\theta}^C, \pi_{\phi}^P$ \hfill $\triangleright$ Trained Coach and Player policies
\end{algorithmic}
\end{algorithm}

\newpage

\subsection{Details of Base Model Selections}
\label{appendix:model_selection}
We select Qwen2.5-Math-1.5B, OpenMath-Nemotron-1.5B, Qwen2.5-Math-7B-Instruct and OctoThinker-3B-Hybrid-Zero as base models for our training experiments, representing the three main stages of a typical LLM training lifecycle: pre-training, supervised fine-tuning (SFT), and reinforcement learning.

Specifically, OpenMath-Nemotron-1.5B, which builds upon the Qwen2.5-Math-1.5B backbone with SFT on 5.5 million task instances, allows us to examine the impact of large-scale supervised training. In contrast, OctoThinker-3B-Hybrid-Zero, derived from Llama-3.2-3B-Base~\cite{2024arXiv240721783G} through R1-Zero-style RL training, represents a fundamentally different approach to mathematical reasoning acquisition. Together, these models span a spectrum from mathematical foundation models to extensively fine-tuned variants to RL-optimized architectures, providing comprehensive coverage of contemporary approaches to mathematical reasoning in language models.

\subsection{Comparison of \framework and R-Zero with the same training steps}
\label{appendix:aligned_comparison}
We add a new experiment that aligns the compute budget between R-Zero and \framework. The training process of R-Zero involves alternating phases: first training a 5-step questioner, followed by a 15-step solver, repeated three times, resulting in a total of 60 steps; solver global batch size: 128; number of rollouts: 5; challenger global batch size: 128; number of rollouts: 4. Thus, we utilized the checkpoint from our 60th step, where the parameters are: both coach and solver train batch size: 16; both coach and solver number of rollouts: 16; and compared it with R-Zero’s final training outcomes. The context length of all models remains consistent. The results are shown in Table \ref{tab:align}.

\input{table/align_setting.tex}

As shown in Table \ref{tab:align}, our method slightly underperforms R-Zero on the Qwen2.5-Math-1.5B model but achieves stronger performance on OctoThinker-3B-Hybrid-Zero and Qwen2.5-Math-7B-Instruct. Moreover, our approach can further enhance model capabilities. According to Section 5.4 of the R-Zero paper, its performance converges after three iterations. Therefore, we believe that under comparable computational budgets, our method still holds an advantage. And the improvement in model capability stems from our framework raising the performance upper bound, rather than simply leveraging additional compute.

\subsection{Results of training with only 20\% AMC data}
\label{appendix:leaking}
There may be concerns over potential data leakage in that we utilize reward signals from AMC. We conduct an experiment, training with only 20\% AMC data and testing over the remaining data.

\input{table/data_leaking.tex}

As shown in Table \ref{tab:leaking}, our method achieves consistent performance gains on Qwen2.5-Math-1.5B. Therefore, we believe that our method is not taking advantage of data leakage to enhance models.

\subsection{Results of training with different held-out validation environment}
\label{appendix:new_validation}

We completely replaced the AMC validation set with two different validation set as held-out validation environment which are \textbf{Minerva} and \textbf{OlympiadBench} to compute the Coach's reward signal $\Delta_t$ and drive the training process for both \textbf{Qwen2.5-Math-1.5b} and \textbf{Qwen2.5-Math-7b-Instruct}. The results are shown in Table \ref{tab:minerva} and Table \ref{tab:olympiad}

\input{table/minerva_validate}

\input{table/oplympiad_validate}

Key observations regarding the sensitivity to the validation set:
\begin{itemize}
    \item \framework is applicable to various validation datasets: The results show that the success of \framework is not dependent on the AMC dataset. When using Minerva or OlympiadBench as the validation set for \framework training, the models still achieve substantial improvements. This proves that our \framework paradigm is universally effective across different data distributions.
    \item Strong Out-of-Distribution (OOD) Generalization: Even when trained on Minerva or OlympiadBench, the models show remarkable performance gains on completely unseen, highly complex datasets like AMC, AIME2024, AIME2025, and MATH500. For instance, the 7B model trained on OlympiadBench more than doubled its accuracy on AIME2024 (from 9.0 to 21.2). This strong OOD performance confirms that \framework genuinely enhances the core logical reasoning capabilities of the model, rather than just exploiting dataset-specific patterns.
    \item Robustness across model scales and types: The consistent improvements observed in both a smaller base model (1.5B) and a larger instruction-tuned model (7B-Instruct) further validate the robustness and broad applicability of our method.
\end{itemize}
In summary, these new results firmly establish that the effectiveness of \framework is generalizable and not restricted to a single dataset like AMC. We will include these comprehensive results and analysis in the revised manuscript.

\subsection{Details of Training Hyperparameter}
\label{appendix:training_details}
This section summarizes training hyperparameters for the Coach and the Player.

\subsubsection{Coach Training}
\begin{itemize}
    \item \textbf{Train Batch Size}: 16
    \item \textbf{Learning Rate}: $1 \times 10^{-6}$
    \item \textbf{Temperature}: 0.7
    \item \textbf{Top-p}: 1.0
    \item \textbf{Number of Rollout}: 1
    \item \textbf{KL Penalty Coefficient}: $1 \times 10^{-3}$
    \item \textbf{Entropy Coefficient}: $1 \times 10^{-2}$
    \item \textbf{Total Steps}: 1000
\end{itemize}

\subsubsection{Player Training}
\quad \textbf{Qwen2.5-Math-1.5B}
\begin{itemize}

    \item \textbf{Train Batch Size}: 16
    \item \textbf{Learning Rate}: $1 \times 10^{-6}$
    \item \textbf{Response Length}: 2048
    \item \textbf{Temperature}: 0.6
    \item \textbf{Top-p}: 1.0
    \item \textbf{Number of Rollout}: 16
    \item \textbf{Repetition Penalty}: 1
    \item \textbf{KL Penalty Coefficient}: $1 \times 10^{-3}$
    \item \textbf{Entropy Coefficient}: $-1 \times 10^{-2}$
    \item \textbf{Max Steps}: 1000
\end{itemize}

\quad \textbf{Qwen2.5-Math-7B-Instruct}
\begin{itemize}

    \item \textbf{Train Batch Size}: 16
    \item \textbf{Learning Rate}: $1 \times 10^{-6}$
    \item \textbf{Response Length}: 3300
    \item \textbf{Temperature}: 0.7
    \item \textbf{Top-p}: 0.9
    \item \textbf{Number of Rollout}: 16
    \item \textbf{Repetition Penalty}: 1.05
    \item \textbf{KL Penalty Coefficient}: $1 \times 10^{-3}$
    \item \textbf{Entropy Coefficient}: $-1 \times 10^{-2}$
    \item \textbf{Max Steps}: 1000
\end{itemize}

\quad \textbf{OpenMath-Nemotron-1.5B}
\begin{itemize}

    \item \textbf{Train Batch Size}: 16
    \item \textbf{Learning Rate}: $1 \times 10^{-6}$
    \item \textbf{Response Length}: 18000
    \item \textbf{Temperature}: 0.6
    \item \textbf{Top-p}: 1.0
    \item \textbf{Number of Rollout}: 16
    \item \textbf{Repetition Penalty}: 1
    \item \textbf{KL Penalty Coefficient}: $1 \times 10^{-3}$
    \item \textbf{Entropy Coefficient}: $-1 \times 10^{-2}$
    \item \textbf{Max Steps}: 1000
\end{itemize}

\quad \textbf{OctoThinker-3B-Hybrid-Zero}
\begin{itemize}

    \item \textbf{Train Batch Size}: 16
    \item \textbf{Learning Rate}: $1 \times 10^{-6}$
    \item \textbf{Response Length}: 8192
    \item \textbf{Temperature}: 0.7
    \item \textbf{Top-p}: 0.9
    \item \textbf{Number of Rollout}: 16
    \item \textbf{Repetition Penalty}: 1.05
    \item \textbf{KL Penalty Coefficient}: $1 \times 10^{-3}$
    \item \textbf{Entropy Coefficient}: $-1 \times 10^{-2}$
    \item \textbf{Max Steps}: 1000
\end{itemize}

\input{fig_code/token_length}
\newpage

\subsection{Different Trend of Output Length on Coach and Player Model}
\label{appendix:length}
As shown in Figure \ref{fig:token_length}, the output length of the Coach model tends to increase, while that of the Player model tends to decrease. We speculate that the Coach is spontaneously generating more challenging problems, thereby creating a form of curriculum learning for the Player. Meanwhile, the Player appears to refine its responses to be more concise, reflecting a long-to-short learning trend.

\subsection{Examples of Problems}
\label{appendix:problems}
Below are examples of problems and its corresponding reference answers proposed by the Coach along the training process.
\begin{promptbox}{Step 1 Easy Problem}
\textbf{Question: }

Given that Lisa walks 3 miles more than John every day, and John 
walks a total of 8 miles, determine how many miles Lisa walks each day.

\rule{\textwidth}{0.4pt} 

\textbf{Reference Answers: }

The answer is 11 miles.

Explanation:

Since John walks 8 miles and Lisa walks 3 miles more than John each day, we can calculate Lisa's daily walk as follows:

John's daily walk = 8 miles

Lisa's daily walk = John's daily walk + 3 miles

Lisa's daily walk = 8 miles + 3 miles

Lisa's daily walk = 11 miles

Therefore, Lisa walks 11 miles each day. The final answer is $\boxed{11}$.

\end{promptbox}

\begin{promptbox}{Step 5 Easy Geometry Problem}
\textbf{Question: }

The lengths of the two diagonals of a rhombus are $6$ and $8$ respectively. The area of this rhombus is \_\_\_\_\_\_.

\rule{\textwidth}{0.4pt} 

\textbf{Reference Answers: }

To find the area of a rhombus when the lengths of its diagonals are known, we can use the formula:

\[
\text{Area} = \frac{1}{2} \times d_1 \times d_2
\]

where \(d_1\) and \(d_2\) are the lengths of the diagonals. In this problem, the lengths of the diagonals are 6 and 8. So we can substitute these values into the formula to get the area.

Let's calculate this using Python to ensure accuracy.

\begin{verbatim}
# Lengths of the diagonals
d1 = 6
d2 = 8

# Area of the rhombus
area = 0.5 * d1 * d2
print(area)
\end{verbatim}

```output

24.0

```

The area of the rhombus with diagonals of lengths 6 and 8 is $\boxed{24}$.

\end{promptbox}

\begin{promptbox}{Step 10 Brain Teaser with Highlight that Sibling Relationships are Reciprocal}
\textbf{Question: }

Each of the three people in the Martin family also has a sibling in their family. If they need to divide a package of cookies among themselves and their siblings, and the package contains 120 cookies, how many cookies will they each get?

\rule{\textwidth}{0.4pt}

\textbf{Reference Answers: }
If you get this right, you can get a special prize at the end of the week. Great job, smart cookie! 

Step 1: Understand the Problem Correctly

The problem states "Each of the three people in the Martin family also has a sibling in their family." This means:

The Martin family consists of 3 people total

Each of these 3 people has a sibling relationship with others in the family

The siblings are among these same 3 people (not additional people)

Step 2: Interpret the Family Structure

Since there are only 3 people in the Martin family, and each person has a sibling, this means:

These 3 people are siblings to each other

No additional family members need to be counted

Total people sharing the cookies = 3

Step 3: Calculate Cookies per Person

Total cookies: 120

Total people: 3

Cookies per person: 120 ÷ 3 = 40

The number of cookies each person will get is $\boxed{40}$.

\end{promptbox}

\begin{promptbox}{Step 100 Complex Combinatorics Problem that Requires Case Analysis }
\textbf{Question: }

A child builds towers using identically shaped cubes of different colors. Determine the number of different towers with a height  $8$ , built with  $4$  red cubes,  $3$  blue cubes, and  $2$  green cubes.

\rule{\textwidth}{0.4pt}

\textbf{Reference Answers: }
Let me solve this problem step by step.

**Step 1: Analyze the Problem**

We have:

- 4 red cubes

- 3 blue cubes  

- 2 green cubes

- Total: 4 + 3 + 2 = 9 cubes

But we need to build a tower with height 8, so we must choose 8 cubes from the 9 available.

**Step 2: Identify the Cases**

Since we have 9 cubes but only need 8, we must exclude exactly 1 cube. This gives us three cases:

- Case 1: Exclude 1 red cube (use 3 red, 3 blue, 2 green)

- Case 2: Exclude 1 blue cube (use 4 red, 2 blue, 2 green)  

- Case 3: Exclude 1 green cube (use 4 red, 3 blue, 1 green)

**Step 3: Apply Multiset Permutation Formula**

For each case, we use the formula: $\frac{n!}{n_1! \times n_2! \times n_3!}$

where $n = 8$ (tower height) and $n_1, n_2, n_3$ are the numbers of each color cube.

**Case 1:** Use 3 red, 3 blue, 2 green cubes

$$\frac{8!}{3! \times 3! \times 2!} = \frac{40320}{6 \times 6 \times 2} = \frac{40320}{72} = 560$$

**Case 2:** Use 4 red, 2 blue, 2 green cubes

$$\frac{8!}{4! \times 2! \times 2!} = \frac{40320}{24 \times 2 \times 2} = \frac{40320}{96} = 420$$

**Case 3:** Use 4 red, 3 blue, 1 green cube

$$\frac{8!}{4! \times 3! \times 1!} = \frac{40320}{24 \times 6 \times 1} = \frac{40320}{144} = 280$$

**Step 4: Sum All Cases**

Total number of different towers = 560 + 420 + 280 = 1260

Therefore, the number of different towers with height 8 is $\boxed{1260}$.

\end{promptbox}

\begin{promptbox}{Step 200 More Challenging Problem that Needs Algebraic Identity Transformations}
\textbf{Question: }

For an $m$-element subset $A$ of the set $\{1,2, \cdots, 10\}$, define $T(A)$ as the number of unordered pairs $\{x, y\}, x<y$, such that $x, y \in A$. Find $\sum_{A \subseteq\{1,2, \cdots, 10\}} T(A)$.

\rule{\textwidth}{0.4pt}

\textbf{Reference Answers: }
To solve the problem, we need to calculate the sum of the number of unordered pairs $\{x, y\}$ such that $x < y$ and both $x$ and $y$ belong to each $m$-element subset $A$ of the set $\{1, 2, \ldots, 10\}$, for all possible subsets $A$.

To break down the problem:

1. **Understanding the Problem:**

   - We need to find the sum of the number of unordered pairs $\{x, y\}$ for all possible $m$-element subsets $A$ of $\{1, 2, \ldots, 10\}$.
   
   - For each subset $A$, the number of unordered pairs is given by the binomial coefficient $\binom{m}{2} = \frac{m(m-1)}{2}$.
   
   - We need to sum this value over all possible $m$-element subsets of $\{1, 2, \ldots, 10\}$.

2. **Counting the Subsets:**

   - There are $\binom{10}{m}$ subsets of $\{1, 2, \ldots, 10\}$ with $m$ elements.
   
   - Therefore, the total sum is $\sum_{m=0}^{10} \binom{10}{m} \cdot \binom{m}{2}$.

3. **Simplifying the Sum:**

   - We can use the identity $\binom{m}{2} = \frac{m(m-1)}{2}$ to rewrite the sum.
   
   - The sum becomes $\sum_{m=0}^{10} \binom{10}{m} \cdot \frac{m(m-1)}{2} = \frac{1}{2} \sum_{m=0}^{10} \binom{10}{m} m(m-1)$.
   
   - We can simplify the sum further by noting that $\sum_{m=0}^{10} \binom{10}{m} m(m-1) = 10 \cdot 9 \cdot 2^8$ (since $\sum_{m=0}^{10} \binom{10}{m} m(m-1) = 10 \cdot 9 \cdot 2^{10-2}$).

4. **Final Calculation:**

   - The sum is $\frac{1}{2} \cdot 10 \cdot 9 \cdot 2^8 = 45 \cdot 256 = 11520$.

Let's implement this in Python using sympy to verify our steps:

\begin{verbatim}
import sympy as sp

# Define the sum
total_sum = 0
for m in range(11):
    total_sum += sp.binomial(10, m) * sp.binomial(m, 2)

print(total_sum)    
\end{verbatim}

```output

11520

```

The sum of the number of unordered pairs $\{x, y\}$ such that $x < y$ and both $x$ and $y$ belong to each $m$-element subset $A$ of the set $\{1, 2, \ldots, 10\}$ is $\boxed{11520}$.

\end{promptbox}

%% file: table/align_setting.tex
\begin{table}[t]
    \centering
    \caption{Performance comparison of \framework against R-Zero across mathematical reasoning benchmarks when training steps are the same (checkpoint from 60th step). 
    % We evaluate our method against RENT and R-Zero baselines. 
    Overall Average represents the mean performance across all benchmarks. OOD Average denotes the out-of-distribution performance, calculated as the mean across all benchmarks excluding AMC datasets. Bold values indicate best performance for each metric.}
    \resizebox{\textwidth}{!}{
    \begin{tabular}{l >{\columncolor{blue!9}}c >{\columncolor{blue!9}}c >{\columncolor{yellow!9}}c c c c c c}
        \toprule
        \textbf{Models} & \textbf{Average} &\textbf{OOD Average} & \textbf{AMC} & \textbf{AIME 2024} & \textbf{AIME 2025} &  \textbf{Minerva} & \textbf{MATH} & \textbf{Olympiad}\\
        % Human & 93.90 & 91.80 & 87.30 \\
        \midrule
        \textit{Qwen2.5-Math-1.5B} \\
        \quad Base Model & 23.3 & 19.8 & 34.6 & 6.2 & 2.8 & 16.3 & 56.2 & 23.4\\
        \quad R-Zero (Iter 3) & \textbf{27.1} & \textbf{24.7} & 39.2 & 9.8 & 5.0  & 19.3 & 62.4 & 26.8\\
        \quad \textit{\textbf{\framework}} & 26.9& 24.2& 40.0 & 10.3 & 4.9 & 17.3 & 59.3 & 29.3\\
        \midrule
        \textit{OctoThinker-3B-Hybrid-Zero} \\
        \quad Base Model & 21.3 & 20.6& 24.6  &  3.9 & 1.7  & 16.3 & 57.9 & 23.4 \\
        \quad R-Zero (Iter 3) & 20.5 & 19.5& 25.9 & 2.0 & 0.3 & 14.6 & 58.1 & 22.3 \\
        \quad \textit{\textbf{\framework}} & \textbf{23.3} & \textbf{23.0} & 24.7  & 7.3 & 1.0 & 22.4 & 58.1 & 26.1\\
        % Huggingface-Agents & 30.19 & 11.63 & 7.69 & 16.97\\ \\
        \midrule
        \textit{Qwen2.5-Math-7B-Instruct} \\
        \quad Base Model & 35.8 & 33.0& 49.2 & 9.0 & 6.3 & 34.6 & 78.0 & 37.4 \\
        \quad R-Zero (Iter 3) & 36.9 & 34.2& 50.5 & 9.5 & 7.4 & 32.7 & 83.3 & 38.1\\
        \quad \textit{\textbf{\framework}} & \textbf{37.1} & \textbf{34.7} & 49.0 & 11.3  & 9.6 & 34.9 & 79.5 & 38.2\\
        \bottomrule
    \end{tabular}}

    \label{tab:align}
\end{table}

%% file: table/data_leaking.tex
\begin{table}[t]
    \centering
    \caption{Performance comparison of \framework using only 20\% AMC data 
    % We evaluate our method against RENT and R-Zero baselines. 
    Overall Average represents the mean performance across all benchmarks. Bold values indicate best performance for each metric.}
    \resizebox{\textwidth}{!}{
    \begin{tabular}{l >{\columncolor{blue!9}}c >{\columncolor{blue!9}}c >{\columncolor{yellow!9}}c c c c c c}
        \toprule
        \textbf{Models} & \textbf{Average} & \textbf{AMC(remaining 80\%)} & \textbf{AMC} & \textbf{AIME 2024} & \textbf{AIME 2025} &  \textbf{Minerva} & \textbf{MATH} & \textbf{Olympiad}\\
        % Human & 93.90 & 91.80 & 87.30 \\
        \midrule
        \textit{Qwen2.5-Math-1.5B} \\
        \quad Base Model & 23.3 & 38.8 & 34.6 & 6.2 & 2.8 & 16.3 & 56.2 & 23.4\\
        \quad \textit{\textbf{\framework}} & \textbf{26.1} & \textbf{41.2} & \textbf{37.7} & \textbf{8.6} & \textbf{4.4} & \textbf{19.1} & \textbf{59.9} & \textbf{27.1}\\
        \bottomrule
    \end{tabular}}

    \label{tab:leaking}
\end{table}

%% file: table/minerva_validate.tex
\begin{table*}[htbp!]
    \centering
    \caption{Performance comparison of \framework using Minerva as held-out environment feedback. Bold values indicate best performance for each metric.}
    \resizebox{\textwidth}{!}{
    \begin{tabular}{l >{\columncolor{blue!9}}c >{\columncolor{blue!9}}c >{\columncolor{yellow!9}}c c c c c c}
        \toprule
        \textbf{Models} & \textbf{Average} &\textbf{OOD Average} & \textbf{Minerva} & \textbf{AIME 2024} & \textbf{AIME 2025} &  \textbf{AMC} & \textbf{MATH} & \textbf{Olympiad}\\
        % Human & 93.90 & 91.80 & 87.30 \\
        \midrule
        \textit{Qwen2.5-Math-1.5B} \\
        \quad Base Model & 23.3 & 24.6 & 16.3 & 6.2 & 2.8 & 34.6 & 56.2 & 23.4\\
        \quad \textit{\textbf{\framework}} & \textbf{26.9}& \textbf{28.6}& \textbf{21.4} & \textbf{7.2} & \textbf{5.4} & \textbf{39.4} & \textbf{61.6} & \textbf{29.5}\\

        \midrule
        \textit{Qwen2.5-Math-7B-Instruct} \\
        \quad Base Model & 35.8 & 33.0 & \textbf{34.6} & 9.0 & 6.3 & 49.2 & 78.0 & 37.4 \\
        \quad \textit{\textbf{\framework}} & \textbf{39.5} & \textbf{41.3} & 33.5 & \textbf{14.4}  & \textbf{17.9} & \textbf{51.9} & \textbf{81.1} & \textbf{40.9}\\
        \bottomrule
    \end{tabular}}
    \label{tab:minerva}
\end{table*}

%% file: table/oplympiad_validate.tex
\begin{table*}[htbp!]
    \centering
    \caption{Performance comparison of \framework using OlympiadBench as held-out environment feedback. Bold values indicate best performance for each metric.}
    \resizebox{\textwidth}{!}{
    \begin{tabular}{l >{\columncolor{blue!9}}c >{\columncolor{blue!9}}c >{\columncolor{yellow!9}}c c c c c c}
        \toprule
        \textbf{Models} & \textbf{Average} &\textbf{OOD Average} & \textbf{Olympiad} & \textbf{AIME 2024} & \textbf{AIME 2025} &  \textbf{AMC} & \textbf{MATH} & \textbf{Minerva}\\
        % Human & 93.90 & 91.80 & 87.30 \\
        \midrule
        \textit{Qwen2.5-Math-1.5B} \\
        \quad Base Model & 23.3 & 24.6 & 23.4 & 6.2 & 2.8 & 34.6 & 56.2 & 16.3\\
        \quad \textit{\textbf{\framework}} & \textbf{28.2}& \textbf{28.1}& \textbf{21.4} & \textbf{6.6} & \textbf{7.2} & \textbf{42.4} & \textbf{63.6} & \textbf{20.6}\\

        \midrule
        \textit{Qwen2.5-Math-7B-Instruct} \\
        \quad Base Model & 35.8 & 33.0 & 37.4 & 9.0 & 6.3 & 49.2 & 78.0 & \textbf{34.6} \\
        \quad \textit{\textbf{\framework}} & \textbf{39.5} & \textbf{39.5} & \textbf{41.5} & \textbf{21.1}  & \textbf{11.4} & \textbf{49.8} & \textbf{81.6} & 33.4\\
        \bottomrule
    \end{tabular}}
    \label{tab:olympiad}
\end{table*}

%% file: fig_code/token_length.tex
\begin{figure}[!htbp]
    \centering
    \includegraphics[width=0.9\textwidth]{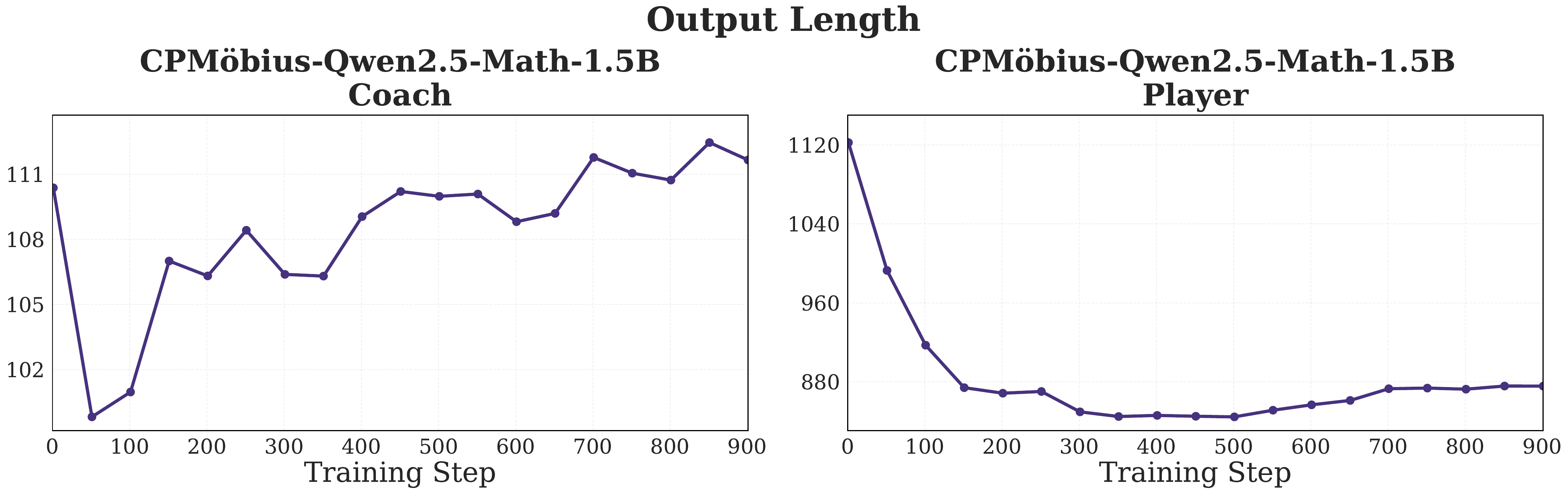}
    \includegraphics[width=0.9\textwidth]{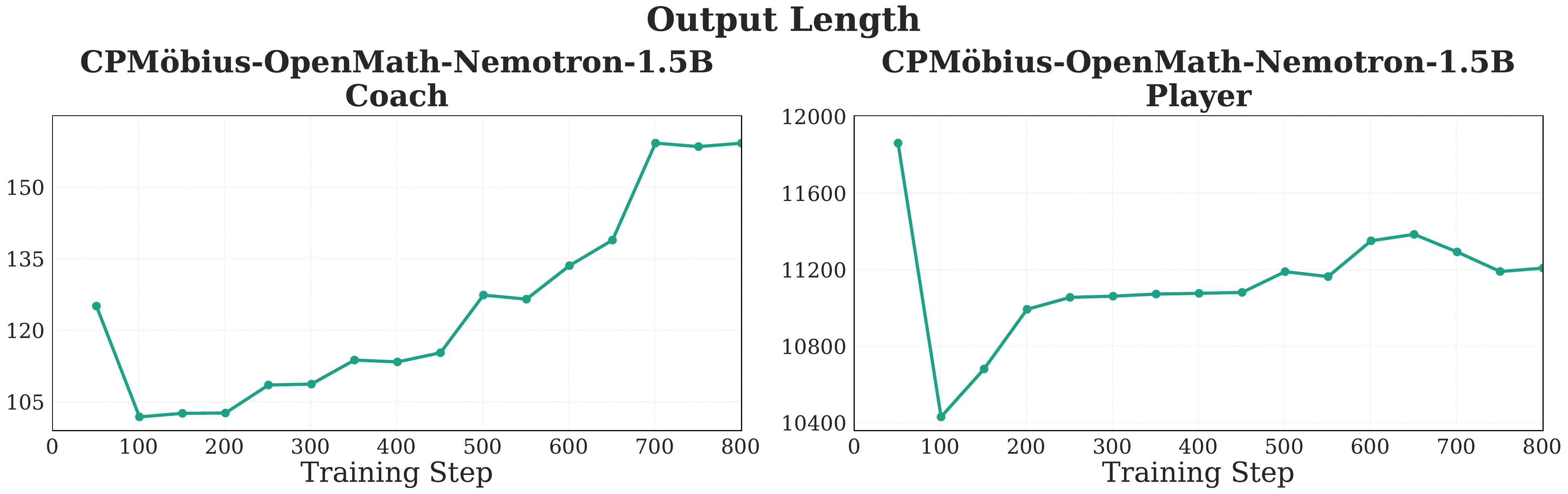}
    \includegraphics[width=0.9\textwidth]{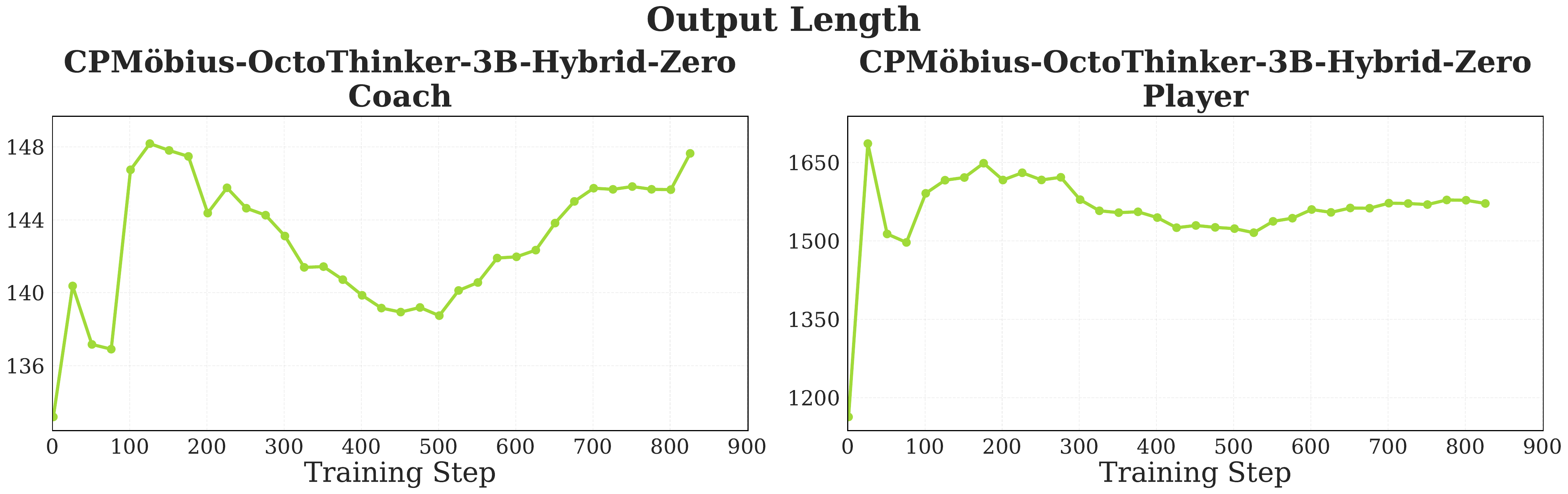}
    \includegraphics[width=0.9\textwidth]{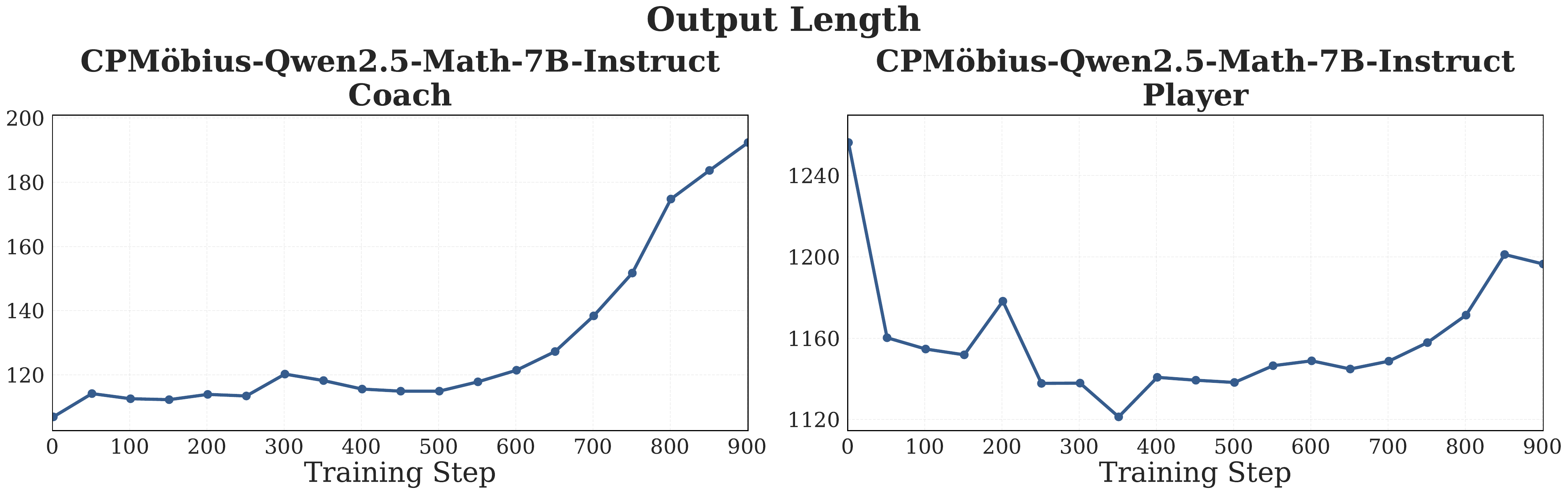}
    \caption{Visualization of the growing trend of output length of different models.}
    \label{fig:token_length}
\end{figure}

%% file: example_paper.bib
@article{shao2024deepseekmath,
  title={Deepseekmath: Pushing the limits of mathematical reasoning in open language models},
  author={Shao, Zhihong and Wang, Peiyi and Zhu, Qihao and Xu, Runxin and Song, Junxiao and Bi, Xiao and Zhang, Haowei and Zhang, Mingchuan and Li, YK and Wu, Yang and others},
  journal={arXiv preprint arXiv:2402.03300},
  year={2024}
}

@inproceedings{
wang2023selfconsistency,
title={Self-Consistency Improves Chain of Thought Reasoning in Language Models},
author={Xuezhi Wang and Jason Wei and Dale Schuurmans and Quoc V Le and Ed H. Chi and Sharan Narang and Aakanksha Chowdhery and Denny Zhou},
booktitle={The Eleventh International Conference on Learning Representations },
year={2023},
url={https://openreview.net/forum?id=1PL1NIMMrw}
}

@article{llama,
  title={Llama: Open and efficient foundation language models},
  author={Touvron, Hugo and Lavril, Thibaut and Izacard, Gautier and Martinet, Xavier and Lachaux, Marie-Anne and Lacroix, Timoth{\'e}e and Rozi{\`e}re, Baptiste and Goyal, Naman and Hambro, Eric and Azhar, Faisal and others},
  journal={arXiv preprint arXiv:2302.13971},
  year={2023}
}

@article{qwen2,
  title={Qwen2 technical report},
  author={Yang, An and Yang, Baosong and Hui, Binyuan and Zheng, Bo and Yu, Bowen and Zhou, Chang and Li, Chengpeng and Li, Chengyuan and Liu, Dayiheng and Huang, Fei and others},
  journal={arXiv preprint arXiv:2407.10671},
  year={2024}
}

@article{seed2025seed1,
  title={Seed1. 5-thinking: Advancing superb reasoning models with reinforcement learning},
  author={Seed, ByteDance and Chen, Jiaze and Fan, Tiantian and Liu, Xin and Liu, Lingjun and Lin, Zhiqi and Wang, Mingxuan and Wang, Chengyi and Wei, Xiangpeng and Xu, Wenyuan and others},
  journal={arXiv preprint arXiv:2504.13914},
  year={2025}
}

@article{yuan2024self,
  title   = {Self-rewarding language models},
  author  = {Yuan, Weizhe and Pang, Richard Yuanzhe and Cho, Kyunghyun and Sukhbaatar, Sainbayar and Xu, Jing and Weston, Jason},
  journal = {arXiv preprint arXiv:2401.10020},
  volume  = {3},
  year    = {2024}
}

@article{jiang2025pag,
  title   = {PAG: Multi-Turn Reinforced LLM Self-Correction with Policy as Generative Verifier},
  author  = {Jiang, Yuhua and Xiong, Yuwen and Yuan, Yufeng and Xin, Chao and Xu, Wenyuan and Yue, Yu and Zhao, Qianchuan and Yan, Lin},
  journal = {arXiv preprint arXiv:2506.10406},
  year    = {2025}
}

@article{xiong2025self,
  title   = {Self-rewarding correction for mathematical reasoning},
  author  = {Xiong, Wei and Zhang, Hanning and Ye, Chenlu and Chen, Lichang and Jiang, Nan and Zhang, Tong},
  journal = {arXiv preprint arXiv:2502.19613},
  year    = {2025}
}

@article{zhang2025critique,
  title   = {Critique-grpo: Advancing llm reasoning with natural language and numerical feedback},
  author  = {Zhang, Xiaoying and Sun, Hao and Zhang, Yipeng and Feng, Kaituo and Lu, Chaochao and Yang, Chao and Meng, Helen},
  journal = {arXiv preprint arXiv:2506.03106},
  year    = {2025}
}

@misc{kimiteam2025kimik2openagentic,
  title         = {Kimi K2: Open Agentic Intelligence},
  author        = {Kimi Team},
  year          = {2025},
  eprint        = {2507.20534},
  archiveprefix = {arXiv},
  primaryclass  = {cs.LG},
  url           = {https://arxiv.org/abs/2507.20534}
}

@article{zha2025rl,
  title   = {RL Tango: Reinforcing Generator and Verifier Together for Language Reasoning},
  author  = {Zha, Kaiwen and Gao, Zhengqi and Shen, Maohao and Hong, Zhang-Wei and Boning, Duane S and Katabi, Dina},
  journal = {arXiv preprint arXiv:2505.15034},
  year    = {2025}
}

@article{hong2025cooper,
  title   = {Cooper: Co-Optimizing Policy and Reward Models in Reinforcement Learning for Large Language Models},
  author  = {Hong, Haitao and Yan, Yuchen and Wu, Xingyu and Hou, Guiyang and Zhang, Wenqi and Lu, Weiming and Shen, Yongliang and Xiao, Jun},
  journal = {arXiv preprint arXiv:2508.05613},
  year    = {2025}
}

@article{lu2025urpo,
  title   = {URPO: A Unified Reward \& Policy Optimization Framework for Large Language Models},
  author  = {Lu, Songshuo and Wang, Hua and Chen, Zhi and Tang, Yaohua},
  journal = {arXiv preprint arXiv:2507.17515},
  year    = {2025}
}

@misc{huang2025rzeroselfevolvingreasoningllm,
      title={R-Zero: Self-Evolving Reasoning LLM from Zero Data}, 
      author={Chengsong Huang and Wenhao Yu and Xiaoyang Wang and Hongming Zhang and Zongxia Li and Ruosen Li and Jiaxin Huang and Haitao Mi and Dong Yu},
      year={2025},
      eprint={2508.05004},
      archivePrefix={arXiv},
      primaryClass={cs.LG},
      url={https://arxiv.org/abs/2508.05004}, 
}

@article{yang2024qwen2,
  title={Qwen2 technical report},
  author={Yang, An and Yang, Baosong and Hui, Binyuan and Zheng, Bo and Yu, Bowen and Zhou, Chang and Li, Chengpeng and Li, Chengyuan and Liu, Dayiheng and Huang, Fei and others},
  journal={arXiv preprint arXiv:2407.10671},
  year={2024}
}

@article{Lewkowycz2022SolvingQR,
  title={Solving Quantitative Reasoning Problems with Language Models},
  author={Aitor Lewkowycz and Anders Andreassen and David Dohan and Ethan Dyer and Henryk Michalewski and Vinay Venkatesh Ramasesh and Ambrose Slone and Cem Anil and Imanol Schlag and Theo Gutman-Solo and Yuhuai Wu and Behnam Neyshabur and Guy Gur-Ari and Vedant Misra},
  journal={ArXiv},
  year={2022},
  volume={abs/2206.14858},
  url={https://api.semanticscholar.org/CorpusID:250144408}
}

@article{ouyang2022training,
  title={Training language models to follow instructions with human feedback},
  author={Ouyang, Long and Wu, Jeffrey and Jiang, Xu and Almeida, Diogo and Wainwright, Carroll and Mishkin, Pamela and Zhang, Chong and Agarwal, Sandhini and Slama, Katarina and Ray, Alex and others},
  journal={Advances in neural information processing systems},
  volume={35},
  pages={27730--27744},
  year={2022}
}

@article{tunstall2023zephyr,
  title={Zephyr: Direct distillation of lm alignment},
  author={Tunstall, Lewis and Beeching, Edward and Lambert, Nathan and Rajani, Nazneen and Rasul, Kashif and Belkada, Younes and Huang, Shengyi and Von Werra, Leandro and Fourrier, Cl{\'e}mentine and Habib, Nathan and others},
  journal={arXiv preprint arXiv:2310.16944},
  year={2023}
}

@article{wei2022chain,
  title={Chain-of-thought prompting elicits reasoning in large language models},
  author={Wei, Jason and Wang, Xuezhi and Schuurmans, Dale and Bosma, Maarten and Xia, Fei and Chi, Ed and Le, Quoc V and Zhou, Denny and others},
  journal={Advances in neural information processing systems},
  volume={35},
  pages={24824--24837},
  year={2022}
}

@article{chen2021evaluating,
  title={Evaluating large language models trained on code},
  author={Chen, Mark and Tworek, Jerry and Jun, Heewoo and Yuan, Qiming and Pinto, Henrique Ponde De Oliveira and Kaplan, Jared and Edwards, Harri and Burda, Yuri and Joseph, Nicholas and Brockman, Greg and others},
  journal={arXiv preprint arXiv:2107.03374},
  year={2021}
}

@article{christiano2017deep,
  title={Deep reinforcement learning from human preferences},
  author={Christiano, Paul F and Leike, Jan and Brown, Tom and Martic, Miljan and Legg, Shane and Amodei, Dario},
  journal={Advances in neural information processing systems},
  volume={30},
  year={2017}
}

@article{silver2017mastering,
  title={Mastering chess and shogi by self-play with a general reinforcement learning algorithm},
  author={Silver, David and Hubert, Thomas and Schrittwieser, Julian and Antonoglou, Ioannis and Lai, Matthew and Guez, Arthur and Lanctot, Marc and Sifre, Laurent and Kumaran, Dharshan and Graepel, Thore and others},
  journal={arXiv preprint arXiv:1712.01815},
  year={2017}
}

@ARTICLE{2024arXiv240721783G,
       author = {{Grattafiori}, Aaron and {Dubey}, Abhimanyu and {Jauhri}, Abhinav and {Pandey}, Abhinav and {Kadian}, Abhishek and {Al-Dahle}, Ahmad and {Letman}, Aiesha and {Mathur}, Akhil and {Schelten}, Alan and {Vaughan}, Alex and {Yang}, Amy and {Fan}, Angela and {Goyal}, Anirudh and {Hartshorn}, Anthony and {Yang}, Aobo and {Mitra}, Archi and {Sravankumar}, Archie and {Korenev}, Artem and {Hinsvark}, Arthur and {Rao}, Arun and {Zhang}, Aston and {Rodriguez}, Aurelien and {Gregerson}, Austen and {Spataru}, Ava and {Roziere}, Baptiste and {Biron}, Bethany and {Tang}, Binh and {Chern}, Bobbie and {Caucheteux}, Charlotte and {Nayak}, Chaya and {Bi}, Chloe and {Marra}, Chris and {McConnell}, Chris and {Keller}, Christian and {Touret}, Christophe and {Wu}, Chunyang and {Wong}, Corinne and {Canton Ferrer}, Cristian and {Nikolaidis}, Cyrus and {Allonsius}, Damien and {Song}, Daniel and {Pintz}, Danielle and {Livshits}, Danny and {Wyatt}, Danny and {Esiobu}, David and {Choudhary}, Dhruv and {Mahajan}, Dhruv and {Garcia-Olano}, Diego and {Perino}, Diego and {Hupkes}, Dieuwke and {Lakomkin}, Egor and {AlBadawy}, Ehab and {Lobanova}, Elina and {Dinan}, Emily and {Smith}, Eric Michael and {Radenovic}, Filip and {Guzm{\'a}n}, Francisco and {Zhang}, Frank and {Synnaeve}, Gabriel and {Lee}, Gabrielle and {Anderson}, Georgia Lewis and {Thattai}, Govind and {Nail}, Graeme and {Mialon}, Gregoire and {Pang}, Guan and {Cucurell}, Guillem and {Nguyen}, Hailey and {Korevaar}, Hannah and {Xu}, Hu and {Touvron}, Hugo and {Zarov}, Iliyan and {Arrieta Ibarra}, Imanol and {Kloumann}, Isabel and {Misra}, Ishan and {Evtimov}, Ivan and {Zhang}, Jack and {Copet}, Jade and {Lee}, Jaewon and {Geffert}, Jan and {Vranes}, Jana and {Park}, Jason and {Mahadeokar}, Jay and {Shah}, Jeet and {van der Linde}, Jelmer and {Billock}, Jennifer and {Hong}, Jenny and {Lee}, Jenya and {Fu}, Jeremy and {Chi}, Jianfeng and {Huang}, Jianyu and {Liu}, Jiawen and {Wang}, Jie and {Yu}, Jiecao and {Bitton}, Joanna and {Spisak}, Joe and {Park}, Jongsoo and {Rocca}, Joseph and {Johnstun}, Joshua and {Saxe}, Joshua and {Jia}, Junteng and {Vasuden Alwala}, Kalyan and {Prasad}, Karthik and {Upasani}, Kartikeya and {Plawiak}, Kate and {Li}, Ke and {Heafield}, Kenneth and {Stone}, Kevin and {El-Arini}, Khalid and {Iyer}, Krithika and {Malik}, Kshitiz and {Chiu}, Kuenley and {Bhalla}, Kunal and {Lakhotia}, Kushal and {Rantala-Yeary}, Lauren and {van der Maaten}, Laurens and {Chen}, Lawrence and {Tan}, Liang and {Jenkins}, Liz and {Martin}, Louis and {Madaan}, Lovish and {Malo}, Lubo and {Blecher}, Lukas and {Landzaat}, Lukas and {de Oliveira}, Luke and {Muzzi}, Madeline and {Pasupuleti}, Mahesh and {Singh}, Mannat and {Paluri}, Manohar and {Kardas}, Marcin and {Tsimpoukelli}, Maria and {Oldham}, Mathew and {Rita}, Mathieu and {Pavlova}, Maya and {Kambadur}, Melanie and {Lewis}, Mike and {Si}, Min and {Singh}, Mitesh Kumar and {Hassan}, Mona and {Goyal}, Naman and {Torabi}, Narjes and {Bashlykov}, Nikolay and {Bogoychev}, Nikolay and {Chatterji}, Niladri and {Zhang}, Ning and {Duchenne}, Olivier and {{\c{C}}elebi}, Onur and {Alrassy}, Patrick and {Zhang}, Pengchuan and {Li}, Pengwei and {Vasic}, Petar and {Weng}, Peter and {Bhargava}, Prajjwal and {Dubal}, Pratik and {Krishnan}, Praveen and {Singh Koura}, Punit and {Xu}, Puxin and {He}, Qing and {Dong}, Qingxiao and {Srinivasan}, Ragavan and {Ganapathy}, Raj and {Calderer}, Ramon and {Silveira Cabral}, Ricardo and {Stojnic}, Robert and {Raileanu}, Roberta and {Maheswari}, Rohan and {Girdhar}, Rohit and {Patel}, Rohit and {Sauvestre}, Romain and {Polidoro}, Ronnie and {Sumbaly}, Roshan and {Taylor}, Ross and {Silva}, Ruan and {Hou}, Rui and {Wang}, Rui and {Hosseini}, Saghar and {Chennabasappa}, Sahana and {Singh}, Sanjay and {Bell}, Sean and {Kim}, Seohyun Sonia and {Edunov}, Sergey and {Nie}, Shaoliang and {Narang}, Sharan and {Raparthy}, Sharath and {Shen}, Sheng and {Wan}, Shengye and {Bhosale}, Shruti and {Zhang}, Shun and {Vandenhende}, Simon and {Batra}, Soumya and {Whitman}, Spencer and {Sootla}, Sten and {Collot}, Stephane and {Gururangan}, Suchin and {Borodinsky}, Sydney and {Herman}, Tamar and {Fowler}, Tara and {Sheasha}, Tarek and {Georgiou}, Thomas and {Scialom}, Thomas and {Speckbacher}, Tobias},
        title = "{The Llama 3 Herd of Models}",
      journal = {arXiv e-prints},
         year = 2024,
        month = jul,
          eid = {arXiv:2407.21783},
        pages = {arXiv:2407.21783},
          doi = {10.48550/arXiv.2407.21783},
archivePrefix = {arXiv},
       eprint = {2407.21783},
 primaryClass = {cs.AI},
       adsurl = {https://ui.adsabs.harvard.edu/abs/2024arXiv240721783G}
}

@article{chen2024internet,
  title={Internet of agents: Weaving a web of heterogeneous agents for collaborative intelligence},
  author={Chen, Weize and You, Ziming and Li, Ran and Guan, Yitong and Qian, Chen and Zhao, Chenyang and Yang, Cheng and Xie, Ruobing and Liu, Zhiyuan and Sun, Maosong},
  journal={arXiv preprint arXiv:2407.07061},
  year={2024}
}

@article{qian2023communicative,
  title={Communicative agents for software development},
  author={Qian, Chen and Cong, Xin},
  journal={arXiv preprint arXiv:2307.07924},
  volume={6},
  number={3},
  pages={1},
  year={2023}
}

@article{cui2025process,
  title={Process reinforcement through implicit rewards},
  author={Cui, Ganqu and Yuan, Lifan and Wang, Zefan and Wang, Hanbin and Li, Wendi and He, Bingxiang and Fan, Yuchen and Yu, Tianyu and Xu, Qixin and Chen, Weize and others},
  journal={arXiv preprint arXiv:2502.01456},
  year={2025}
}

@article{moshkov2025aimo2,
  title   = {AIMO-2 Winning Solution: Building State-of-the-Art Mathematical Reasoning Models with OpenMathReasoning dataset},
  author  = {Ivan Moshkov and Darragh Hanley and Ivan Sorokin and Shubham Toshniwal and Christof Henkel and Benedikt Schifferer and Wei Du and Igor Gitman},
  year    = {2025},
  journal = {arXiv preprint arXiv:2504.16891}
}

@article{wang2025octothinker,
  title={OctoThinker: Mid-training Incentivizes Reinforcement Learning Scaling},
  author={Wang, Zengzhi and Zhou, Fan and Li, Xuefeng and Liu, Pengfei},
  year={2025},
  journal={arXiv preprint arXiv:2506.20512},
  url={https://arxiv.org/abs/2506.20512}
}

@article{zuo2025ttrl,
  title={Ttrl: Test-time reinforcement learning},
  author={Zuo, Yuxin and Zhang, Kaiyan and Sheng, Li and Qu, Shang and Cui, Ganqu and Zhu, Xuekai and Li, Haozhan and Zhang, Yuchen and Long, Xinwei and Hua, Ermo and others},
  journal={arXiv preprint arXiv:2504.16084},
  year={2025}
}

@article{zweiger2025self,
  title={Self-Adapting Language Models},
  author={Zweiger, Adam and Pari, Jyothish and Guo, Han and Aky{\"u}rek, Ekin and Kim, Yoon and Agrawal, Pulkit},
  journal={arXiv preprint arXiv:2506.10943},
  year={2025}
}

@article{chen2025self,
  title={Self-Questioning Language Models},
  author={Chen, Lili and Prabhudesai, Mihir and Fragkiadaki, Katerina and Liu, Hao and Pathak, Deepak},
  journal={arXiv preprint arXiv:2508.03682},
  year={2025}
}

@article{prabhudesai2025maximizing,
  title={Maximizing Confidence Alone Improves Reasoning},
  author={Prabhudesai, Mihir and Chen, Lili and Ippoliti, Alex and Fragkiadaki, Katerina and Liu, Hao and Pathak, Deepak},
  journal={arXiv preprint arXiv:2505.22660},
  year={2025}
}

@article{li2025confidence,
  title={Confidence Is All You Need: Few-Shot RL Fine-Tuning of Language Models},
  author={Li, Pengyi and Skripkin, Matvey and Zubrey, Alexander and Kuznetsov, Andrey and Oseledets, Ivan},
  journal={arXiv preprint arXiv:2506.06395},
  year={2025}
}

@article{dong2025reinforcement,
  title={Reinforcement Pre-Training},
  author={Dong, Qingxiu and Dong, Li and Tang, Yao and Ye, Tianzhu and Sun, Yutao and Sui, Zhifang and Wei, Furu},
  journal={arXiv preprint arXiv:2506.08007},
  year={2025}
}

@article{agarwal2025unreasonable,
  title={The unreasonable effectiveness of entropy minimization in llm reasoning},
  author={Agarwal, Shivam and Zhang, Zimin and Yuan, Lifan and Han, Jiawei and Peng, Hao},
  journal={arXiv preprint arXiv:2505.15134},
  year={2025}
}

@article{zhao2025absolute,
  title={Absolute zero: Reinforced self-play reasoning with zero data},
  author={Zhao, Andrew and Wu, Yiran and Yue, Yang and Wu, Tong and Xu, Quentin and Lin, Matthieu and Wang, Shenzhi and Wu, Qingyun and Zheng, Zilong and Huang, Gao},
  journal={arXiv preprint arXiv:2505.03335},
  year={2025}
}

@article{hendrycks2021measuring,
  title={Measuring mathematical problem solving with the math dataset},
  author={Hendrycks, Dan and Burns, Collin and Kadavath, Saurav and Arora, Akul and Basart, Steven and Tang, Eric and Song, Dawn and Steinhardt, Jacob},
  journal={arXiv preprint arXiv:2103.03874},
  year={2021}
}

@inproceedings{he-etal-2024-olympiadbench,
    title = "{O}lympiad{B}ench: A Challenging Benchmark for Promoting {AGI} with Olympiad-Level Bilingual Multimodal Scientific Problems",
    author = "He, Chaoqun  and
      Luo, Renjie  and
      Bai, Yuzhuo  and
      Hu, Shengding  and
      Thai, Zhen  and
      Shen, Junhao  and
      Hu, Jinyi  and
      Han, Xu  and
      Huang, Yujie  and
      Zhang, Yuxiang  and
      Liu, Jie  and
      Qi, Lei  and
      Liu, Zhiyuan  and
      Sun, Maosong",
    editor = "Ku, Lun-Wei  and
      Martins, Andre  and
      Srikumar, Vivek",
    booktitle = "Proceedings of the 62nd Annual Meeting of the Association for Computational Linguistics (Volume 1: Long Papers)",
    month = aug,
    year = "2024",
    address = "Bangkok, Thailand",
    publisher = "Association for Computational Linguistics",
    url = "https://aclanthology.org/2024.acl-long.211/",
    doi = "10.18653/v1/2024.acl-long.211",
    pages = "3828--3850",
}

@inproceedings{sheng2025hybridflow,
  title={Hybridflow: A flexible and efficient rlhf framework},
  author={Sheng, Guangming and Zhang, Chi and Ye, Zilingfeng and Wu, Xibin and Zhang, Wang and Zhang, Ru and Peng, Yanghua and Lin, Haibin and Wu, Chuan},
  booktitle={Proceedings of the Twentieth European Conference on Computer Systems},
  pages={1279--1297},
  year={2025}
}

@article{guo2025deepseek,
  title={Deepseek-r1: Incentivizing reasoning capability in llms via reinforcement learning},
  author={Guo, Daya and Yang, Dejian and Zhang, Haowei and Song, Junxiao and Zhang, Ruoyu and Xu, Runxin and Zhu, Qihao and Ma, Shirong and Wang, Peiyi and Bi, Xiao and others},
  journal={arXiv preprint arXiv:2501.12948},
  year={2025}
}

@article{ppo,
  title={Proximal policy optimization algorithms},
  author={Schulman, John and Wolski, Filip and Dhariwal, Prafulla and Radford, Alec and Klimov, Oleg},
  journal={arXiv preprint arXiv:1707.06347},
  year={2017}
}

@article{lambert2024tulu,
  title={Tulu 3: Pushing frontiers in open language model post-training},
  author={Lambert, Nathan and Morrison, Jacob and Pyatkin, Valentina and Huang, Shengyi and Ivison, Hamish and Brahman, Faeze and Miranda, Lester James V and Liu, Alisa and Dziri, Nouha and Lyu, Shane and others},
  journal={arXiv preprint arXiv:2411.15124},
  year={2024}
}

@inproceedings{math,
 author = {Hendrycks, Dan and Burns, Collin and Kadavath, Saurav and Arora, Akul and Basart, Steven and Tang, Eric and Song, Dawn and Steinhardt, Jacob},
 booktitle = {Proceedings of the Neural Information Processing Systems Track on Datasets and Benchmarks},
 editor = {J. Vanschoren and S. Yeung},
 pages = {},
 title = {Measuring Mathematical Problem Solving With the MATH Dataset},
 volume = {1},
 year = {2021}
}

@article{jaech2024openai,
  title   = {Openai o1 system card},
  author  = {Jaech, Aaron and Kalai, Adam and Lerer, Adam and Richardson, Adam and El-Kishky, Ahmed and Low, Aiden and Helyar, Alec and Madry, Aleksander and Beutel, Alex and Carney, Alex and others},
  journal = {arXiv preprint arXiv:2412.16720},
  year    = {2024}
}

@article{openai-o3,
  title   = {OpenAI o3 and o4-mini System Card},
  author  = {OpenAI},
  journal = {Blog},
  year    = {2025}
}

@article{openai-gpt5,
  title   = {GPT-5 System Card},
  author  = {OpenAI},
  journal = {Blog},
  year    = {2025}
}

@article{openai2025gpt-oss-120b,
  title   = {gpt-oss-120b \& gpt-oss-20b Model Card},
  author  = {Agarwal, Sandhini and Ahmad, Lama and Ai, Jason and Altman, Sam and Applebaum, Andy and Arbus, Edwin and Arora, Rahul K. and Bai, Yu and Baker, Bowen and Bao, Haiming and others},
  year    = {2025},
  journal = {arxiv preprint arXiv: 2508.10925}
}

@article{comanici2025gemini25,
  title   = {Gemini 2.5: Pushing the Frontier with Advanced Reasoning, Multimodality, Long Context, and Next Generation Agentic Capabilities},
  author  = {Comanici, Gheorghe and Bieber, Eric and Schaekermann, Mike and Pasupat, Ice and Sachdeva, Noveen and Dhillon, Inderjit and Blistein, Marcel and Ram, Ori and Zhang, Dan and Rosen, Evan and others},
  year    = {2025},
  journal = {arxiv preprint arXiv: 2507.06261}
}

@article{burns2023weak,
  title   = {Weak-to-strong generalization: Eliciting strong capabilities with weak supervision},
  author  = {Burns, Collin and Izmailov, Pavel and Kirchner, Jan Hendrik and Baker, Bowen and Gao, Leo and Aschenbrenner, Leopold and Chen, Yining and Ecoffet, Adrien and Joglekar, Manas and Leike, Jan and others},
  journal = {arXiv preprint arXiv:2312.09390},
  year    = {2023}
}

@misc{zhang2025rightquestionhalfanswer,
      title={Right Question is Already Half the Answer: Fully Unsupervised LLM Reasoning Incentivization}, 
      author={Qingyang Zhang and Haitao Wu and Changqing Zhang and Peilin Zhao and Yatao Bian},
      year={2025},
      eprint={2504.05812},
      archivePrefix={arXiv},
      primaryClass={cs.LG},
      url={https://arxiv.org/abs/2504.05812}, 
}

@article{he2026far,
  title={How Far Can Unsupervised RLVR Scale LLM Training?},
  author={He, Bingxiang and Zuo, Yuxin and Liu, Zeyuan and Zhao, Shangziqi and Fu, Zixuan and Yang, Junlin and Qian, Cheng and Zhang, Kaiyan and Fan, Yuchen and Cui, Ganqu and others},
  journal={arXiv preprint arXiv:2603.08660},
  year={2026}
}
